\documentclass{article}

\usepackage{microtype}
\usepackage{graphicx}
\usepackage{subcaption}
\usepackage{svg}
\usepackage{booktabs} % for professional tables
\usepackage{titletoc}

\usepackage{xcolor}   % for \textcolor
\usepackage{xspace}   % for \xspace
\usepackage{pifont} % for \ding, \cmark, \xmark
\newcommand{\cmark}{\ding{51}\xspace}%
\newcommand{\xmarkg}{\textcolor{lightgray}{\ding{55}}\xspace}%
\newcommand{\pub}[1]{\color{gray}{\scriptsize{[{#1}]}}}
\newcommand{\myparagraph}[1]{{\vspace{.1em} \noindent \bf #1}}
\usepackage{colortbl}

\usepackage{hyperref}

\usepackage[accepted]{icml2026}

\usepackage{amsmath}
\usepackage{amssymb}
\usepackage{mathtools}
\usepackage{amsthm}

\usepackage{enumitem}

\usepackage{float}

\definecolor{lightblue}{RGB}{235,245,255}
\definecolor{lightgray}{gray}{0.93}

\usepackage{tcolorbox}
\tcbuselibrary{listings,breakable}
\usepackage{fvextra}

\usepackage[capitalize,noabbrev]{cleveref}

\theoremstyle{plain}

\theoremstyle{definition}

\theoremstyle{remark}

\usepackage[textsize=tiny]{todonotes}
\presetkeys{todonotes}{inline}{}
\usepackage{cuted}  % for teaser figure
\usepackage{multirow}
\usepackage{pifont}      % 用于 \ding{51} (✓) 和 \ding{55} (✗)

\icmltitlerunning{AVTrack: Audio-Visual Tracking in Human-centric Complex Scenes}

\begin{document}

\twocolumn[
  \icmltitle{AVTrack: Audio-Visual Tracking in Human-centric Complex Scenes}



  \icmlsetsymbol{equal}{*}

  \begin{icmlauthorlist}
    \icmlauthor{Yaoting Wang}{fdu}
    \icmlauthor{Yun Zhou}{fdu}
    \icmlauthor{Zipei Zhang}{fdu}
    \icmlauthor{Henghui Ding}{fdu}
  \end{icmlauthorlist}

  \icmlaffiliation{fdu}{Institute of Big Data, College of Computer Science and Artificial Intelligence, Fudan University, Shanghai, China}
  % \icmlaffiliation{yyy}{Department of XXX, Fudan University, Shanghai, China}

  \icmlcorrespondingauthor{Henghui Ding}{hhding@fudan.edu.cn}

  \icmlkeywords{multimodal, audio-visual, segmentation, tracking, ICML}

  % \vskip 0.2in
]



\printAffiliationsAndNotice{}  % no special notice (required even if empty)

\begin{strip}
    \centering
    \vspace{-16mm}
    \includegraphics[width=1\linewidth]{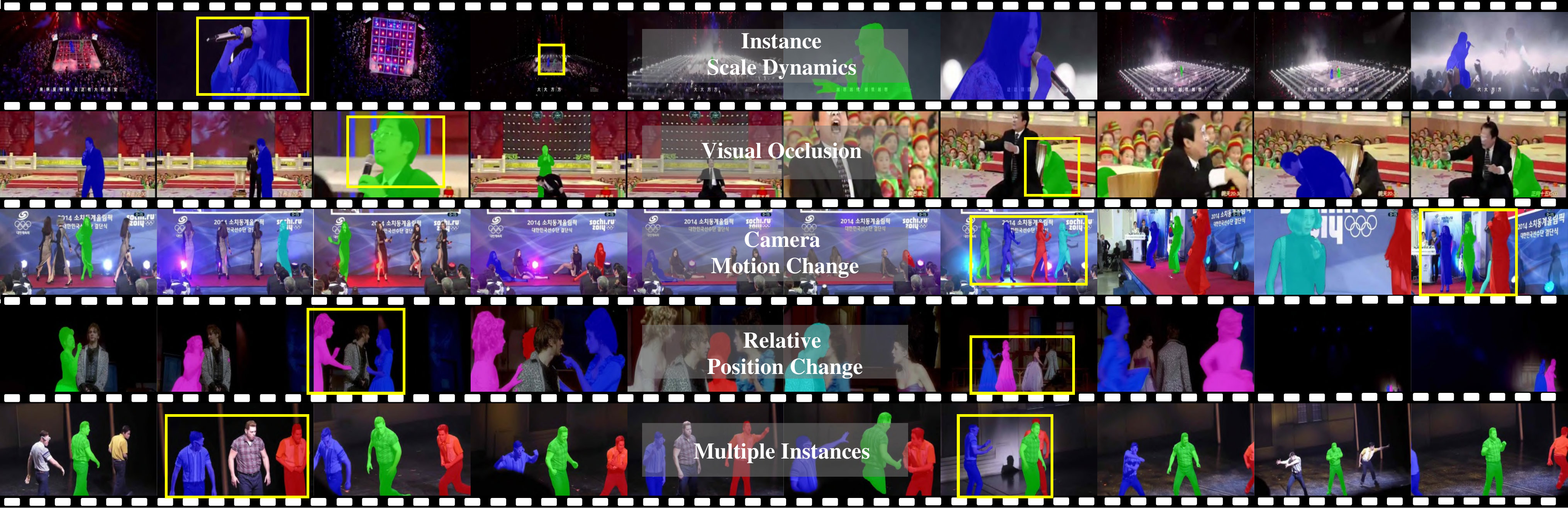}
    \captionof{figure}{Illustrative samples from our proposed AVTrack benchmark. Audio signals are omitted for visual clarity. AVTrack features challenging human-centric audio-visual scenarios, such as instance scale dynamics, visual occlusion, camera motion change, and relative position change. In contrast, previous datasets are typically dominated by simple settings, such as static cameras and single-instance scenes. A more detailed description is provided in \cref{sec:dataset}. Zoom in to inspect the instances in extremely small sizes.
    }
\label{fig:teaser}
\end{strip}

\begin{abstract}
Audio-visual speaker tracking aims to localize and track active speakers by leveraging auditory and visual cues, enabling fine-grained, human-centric scene understanding.
This capability is essential for real-world applications such as intelligent video editing, surveillance, and human–computer interaction.
However, existing datasets are largely limited to simple or homogeneous audio-visual scenes with 
coarse annotations.
Such oversimplified settings bias evaluation toward static audio–visual co-occurrence, rather than rigorously assessing robust spatiotemporal modeling and cross-modal reasoning in complex, dynamic scenes.
To address these limitations, we introduce \textbf{AVTrack}, a human-centric audio-visual instance segmentation (AVIS) dataset designed for dynamic real-world scenarios.
AVTrack features diverse and challenging conditions, including camera motion, visual occlusions, and position changes.
Evaluations of representative AVIS methods on AVTrack reveal substantial performance degradation, establishing AVTrack as a challenging benchmark for robust human-centric audio-visual scene understanding in complex environments.
We further provide a simple yet effective baseline to facilitate future research. Project website: \url{https://FudanCVL.github.io/AVTrack/}.

\end{abstract}

% \newpage

\section{Introduction}
\label{sec:intro}
% \vspace{1mm}
% (why audio-visual) 
% + (audio visual speaker tracking)
% + (audio-visual tracking)
Humans naturally perceive and interpret their surroundings by seamlessly integrating visual and auditory signals. 
This cross-modal capability allows us to effortlessly identify who is speaking in a group, track a clapping individual amid visual clutter, or follow human activities under occlusion and rapid motion. 
Inspired by this perceptual ability, prior work~\cite{li2022multi,zhao2023audio,li2025multi} has explored \emph{audio-visual speaker tracking}, which leverages synchronized audio and visual streams to localize sounding individuals while maintaining identity consistency over time.
As a core component of multimodal perception, audio-visual tracking underpins a wide range of applications, including human--computer interaction, augmented reality, security surveillance, and intelligent video editing.

% \vspace{1.2mm}
Beyond speaker localization, recent progress in \emph{audio-visual segmentation (AVS)}~\cite{senocak2018learning,zhou2022avs,wang2024refavs,guo2025aviseg} has significantly advanced multimodal scene understanding by enabling pixel-level delineation of sounding objects.
Among these efforts, \emph{audio-visual instance segmentation (AVIS)}~\cite{guo2025aviseg} represents a critical step forward, aiming to simultaneously detect, segment, and track sounding instances across video frames at the object level.
By integrating instance-level reasoning with audio-visual correspondence, AVIS holds substantial promise for robust perception in complex, real-world settings, particularly in human-centric scenes.

\vspace{1.2mm}
Building reliable audio-visual correspondence is central to effective fine-grained multimodal learning.
Early works such as AVS-Bench~\cite{zhou2022avs,zhou2025avss} establish end-to-end CNN-based frameworks for jointly modeling visual and auditory cues.
To better capture long-range temporal dependencies, AVSegFormer~\cite{gao2024avsegformer} introduces transformer-based architectures that significantly enhance spatiotemporal audio-visual reasoning.
More recently, GAVS~\cite{wang2024gavs} demonstrates strong data efficiency by integrating audio-visual correlation adapters into the visual foundation model SAM \cite{kirillov2023segment} via adapter tuning.
In parallel, COMBO~\cite{yang2024combo} advances cross-modal fusion by explicitly modeling bilateral audio-visual relations.
Building upon these advances, AVISM~\cite{guo2025aviseg} combines Mask2Former \cite{cheng2022masked} and VITA \cite{heo2022vita} extending audio-visual understanding to the instance level with window-based attention to achieve memory-efficient AVIS.

\vspace{1.2mm}
Despite notable progress, existing tasks and benchmarks still fall short in realistically evaluating AVIS systems. Early audio-visual speaker tracking datasets~\cite{lathoud2004av16,qian2019multi,qian2022audio} are mainly collected in controlled laboratory settings, with few speakers and largely static scenes. Although AVA-ActiveSpeaker~\cite{roth2020ava} increases speaker diversity, it is restricted to a single data source (TV series) and provides only bounding-box annotations without cross-frames identity consistency.
Meanwhile, AVS benchmarks \cite{zhou2022avs,zhou2025avss} offer pixel-level labels in in-the-wild scenarios, but are dominated by short, trimmed clips (5--10 seconds), limiting long-range temporal modeling. AVISeg~\cite{guo2025aviseg} partially addresses this issue by extending clip duration to around 60 seconds for instance-level segmentation; however, most scenes remain visually simple, with limited camera motion, background variation, and relative position shifts, particularly in human-centric scenarios.
Such oversimplified settings fail to capture human-centric real-world challenges such as frequent occlusions, multi-person interactions, and complex spatiotemporal dynamics in the human-centric applications.

\vspace{1.2mm}
To address these challenges, we introduce \textbf{AVTrack}, a novel audio-visual benchmark designed for human-centric AVIS in complex scenes. AVTrack includes 871 video clips for 3,120 densely annotated instance tracklets, spanning 8 challenging conditions such as Camera Motion Change, Visual Occlusions, and Multiple Instances, which significantly complicate the modeling of cross-modal spatiotemporal dynamics. 
We benchmark state-of-the-art Video Instance Segmentation (VIS) and AVIS methods on AVTrack and observe significant performance degradation compared to existing datasets. These findings highlight AVTrack as a more realistic and challenging testbed for evaluating the robustness and advancements of human-centric AVIS systems. 
Additionally, we introduce a simple yet effective baseline framework to further support the community's research in fine-grained and human-centric audio-visual understanding.

In summary, our main contributions are as follows:
\begin{itemize}[
  leftmargin=*,
  topsep=2pt,
  itemsep=4pt,
  parsep=0pt
]
    \item We introduce \textbf{AVTrack} for benchmarking human-centric AVIS in complex and dynamic scenes.
    \item Experiments on AVTrack reveal significant performance gaps for VIS and AVIS methods, and we present a simple yet effective baseline with modular, plug-and-play extensibility for future research.
    \item We perform an in-depth analysis of the challenges posed by AVTrack and discuss promising directions for advancing human-centric AVIS in complex scenes.
\end{itemize}

\vspace{4.5mm}

\section{Related Works}
\label{sec:related}

% VIS的方法VITA是22年的有点老了，加个最近的方法吧；2）在Complex Scenes里面AVISeg就不必再说一遍是扩展到VIS，直接说60 seconds这点；3）MOSE的介绍再短点，现在用了4行有点失衡；3）咱们领域常用一般现在时，通常情况下没必要把时态变来变去

% % https://arxiv.org/pdf/2307.14392
% VIS 末尾 加上 human-centric.

\subsection{Audio-Visual Speaker Tracking}
Audio-visual speaker tracking aims to track active speakers. Prior work has primarily relied on a small set of controlled datasets that limit real-world complexity. The most commonly used classic corporas such as AV16.3 \cite{lathoud2004av16} and CAV3D \cite{qian2019multi} provide multi-speaker audio-visual recordings in indoor meeting settings, but consist of recordings with controlled camera and microphone configurations, constraining diversity in scene dynamics and speaker variation. The AVRI \cite{qian2022audio} dataset increases recording duration, but remains constrained in environment and annotation scope. These benchmarks are valuable for evaluating tracking algorithms but do not capture the complexity of unconstrained real-world audio-visual scenarios beyond laboratory conditions.

\subsection{Audio-Visual Segmentation}
AVS \cite{zhou2022avs} aims to locate and segment sounding objects in a video, serving as a key step toward multimodal scene understanding. The pioneering work AVSBench \cite{zhou2022avs} establishes the foundational benchmark with audio-visual object segmentation for single sound sources (AVS-SS) and for multiple sound sources (AVS-MS), which focus on producing binary masks corresponding to active audio emitters. 
Building upon this, audio-visual semantic segmentation (AVSS) \cite{zhou2025avss} further enriches the task by introducing semantic labeling, requiring the model to predict category-specific segmentation maps aligned with sound cues. Ref-AVS \cite{wang2024refavs} further advances multimodal grounding by introducing the reference audio-visual segmentation (Ref-AVS) task, which leverages textual cues describing visual and auditory information to guide flexible, context-aware segmentation.
Recently, AVISeg \cite{guo2025aviseg} extends the audio-visual domain to AVIS, integrating detection, instance-level mask generation, and temporal association across frames. 
AVS is challenging due to complex cross-modal correspondence and multi-source interference. Existing AVS datasets, however, mostly contain simple and static scenes, limiting their ability to reflect real-world audio-visual complexity.

\subsection{Complex Scene Understanding}
Complex scene understanding has long been a challenge in computer vision, requiring robust reasoning under occlusion, dense interactions, and long-range temporal dependencies. Building on early work in cluttered environments, OVIS~\cite{qi2022ovis} establishes the first large-scale benchmark for occluded video instance segmentation, and MOSE~\cite{ding2023mose} extends it to more realistic settings with dense crowds, camera motion, and large spatial displacements.
Complex scene understanding has also been explored in audio-visual settings.
AVS-MS \cite{zhou2022avs} introduces multi-source audio conditions, requiring the simultaneous segmentation of multiple sounding objects.
More recent works \cite{wang2024teso,zha2025implicit} emphasize the inherent inconsistency between auditory and visual cues, highlighting the need for coherent cross-modal reasoning beyond simple audio-conditioned segmentation.
Despite these advances, existing AVS benchmarks \cite{zhou2022avs,zhou2025avss,wang2024refavs} are largely limited to short clips of 5--10 seconds with simplified spatiotemporal structures, often reducing the task to frame-level audio-conditioned image segmentation.
AVISeg \cite{guo2025aviseg} partially alleviates this limitation by extending video duration to around 60 seconds, yet it still lacks sufficient scene diversity and temporal complexity.
In contrast, our work focuses on human-centric AVIS in 8 complex scenes, aiming to capture dynamic and intricate instance-level audio-visual interactions in real-world settings.

\vspace{2mm}

\section{AVTrack Dataset}
\label{sec:dataset}
\vspace{0.5mm}
The AVTrack dataset is purpose-built to evaluate \emph{human-centric AVIS} under realistic and highly challenging conditions.
Unlike existing benchmarks that prioritize training supervision, AVTrack is released exclusively as a test set.
By decoupling evaluation from dataset-specific training, this design choice reflects our primary objective: to establish a \emph{rigorously curated, annotation-intensive benchmark} that serves as a stable and long-term reference for measuring progress in complex, fine-grained human-centric audio-visual understanding.

\subsection{Define Complex Audio-Visual Scenes}
\label{sec:complex_case}
To move beyond the simple and static scenarios that dominate existing audio-visual instance segmentation (AVIS) datasets, we explicitly define and enforce a set of criteria for \emph{complex audio-visual scenes}.
These criteria are grounded in a systematic analysis of real-world human-centric video content and are specifically designed to capture challenging conditions that expose the limitations of current models.
A video is retained in AVTrack only if it exhibits one or more of the following characteristics:
\begin{itemize}[
  leftmargin=*,
  topsep=2pt,
  itemsep=4pt,
  parsep=0pt
]
    \item \textbf{Visual Occlusion:} Sounding individuals are partially occluded by other objects or people, leading to overlapping instances and ambiguous visual boundaries.
    
    \item \textbf{Relative Position Change:} Spatial ordering of instances changes over time (e.g., a man moves from left to right of others), necessitating persistent identity tracking rather than static localization.
    
    \item \textbf{Background Switch:} The scene undergoes substantial background changes, such as transitions between distinct environments, disrupting visual continuity.
    
    \item \textbf{Camera Motion Change:} Pronounced camera viewpoint variations, including zooming, panning, and shot transitions, alter instance scale, perspective, and visibility.
    
    \item \textbf{Multiple Instances:} Multiple instances of the same semantic class (i.e., humans) appear simultaneously, with only a subset actively producing sound, increasing the difficulty of audio-visual association.
    
    \item \textbf{Multi-turn Sounding:} Speaking turns alternate frequently among different individuals, causing the active audio-visual correspondence to shift over time.
    
    \item \textbf{Audio-Visual Inconsistency:} The auditory signal does not trivially align with the visually salient instance, such as off-screen speakers or background narration.

    % \vspace{15mm}
    \item \textbf{Instance Scale Dynamics:} Sounding individuals may appear at extremely large or small spatial scales and undergo substantial scale variation over time, hindering reliable instance detection and consistent audio-visual correlation.

\end{itemize}

\subsection{Dataset Statistics}

\begin{table*}[tb]
    \centering
    \caption{
Comparison of non-laboratory datasets for audio-visual and visual-only tasks. 
Only VIS and AVIS provide instance-level annotations. 
AVTrack is designed specifically for human-centric AVIS evaluation. 
\textbf{Test}: proportion of test-set; 
\textbf{Length}: average video duration; 
\textbf{Anno.}: annotation granularity; 
\textbf{Audio}: whether audio is provided; 
\textbf{Track}: whether cross-frame instance identity is available.
}
    \label{tab:datasets}
    \resizebox{\textwidth}{!}{%
    \begin{tabular}{clrrrcccccc}
    \toprule
    \textbf{Task} & \textbf{Dataset} &\ \ \textbf{Videos} & \textbf{Test} & \textbf{Length} & \textbf{Domain} & \textbf{Anno.} & \textbf{Audio} & \textbf{Track} & \textbf{Publication} \\
    \midrule
    
    \multirow{1}{*}{\centering ASD} 
        & AVA-ActiveSpeaker \cite{roth2020ava} 
        &  262 & 41.6 & 529.0s & Human & bbox & \cmark & \xmarkg & \pub{ICASSP'20} \\
    % \hline
    \multirow{1}{*}{\centering AVL} 
        & VGG-SS \cite{chen2021localizing} 
        & 5,158 & 100.0 & 10.0s & Common & bbox & \cmark & \xmarkg & \pub{CVPR'21} \\
    % \hline
    
    \multirow{1}{*}{\centering AVOS} 
        & AVSBench-O \cite{zhou2022avs} 
        & 5,356 & 15.0 & 5.0s & Common & mask & \cmark & \xmarkg & \pub{ECCV'22} \\
    % \hline
    \multirow{1}{*}{\centering AVSS} 
        & AVSBench-S \cite{zhou2025avss} 
        & 12,356 & 20.7 & 7.8s & Common & mask & \cmark & \xmarkg & \pub{IJCV'25} \\
    % \hline
    \multirow{1}{*}{\centering Ref-AVS} 
        & RefAVS-Bench \cite{wang2024refavs} 
        & 4,002 & 20.4 & 10.0s & Common & mask & \cmark & \xmarkg & \pub{ECCV'24} \\
    \multirow{1}{*}{\centering Ref-VOS} 
    & J-HMDB Sentences \cite{gavrilyuk2018actor} 
    & 928 & 100.0 & 1.0s & Human & mask & \xmarkg & \xmarkg & \pub{CVPR'18} \\
    \midrule
    \multirow{3}{*}{\centering VIS} 
        & YouTube-VIS \cite{yang2019vis} 
        & 2,883 & 11.9 & 4.6s & Common & mask & \xmarkg & \cmark & \pub{ICCV'19} \\
        & OVIS \cite{qi2022ovis} 
        & 901 & 17.1 & 12.8s & Common & mask & \xmarkg & \cmark & \pub{IJCV'22} \\
        & YouMVOS \cite{wei2022youmvos} 
        & 200 & 15.0 & 333.1s & Human & mask & \xmarkg & \cmark & \pub{CVPR'22} \\
    \midrule
    
    \multirow{2}{*}{\centering AVIS} 
        & AVISeg \cite{guo2025aviseg} 
        & 926 & 22.1 & 61.4s & Common & mask & \cmark & \cmark & \pub{CVPR'25} \\
        
        & \cellcolor{lightblue} \textbf{AVTrack (ours)} 
        & \cellcolor{lightblue}871 & \cellcolor{lightblue}100.0 & \cellcolor{lightblue}54.0s & \cellcolor{lightblue}Human & \cellcolor{lightblue}mask & \cellcolor{lightblue}\cmark & \cellcolor{lightblue}\cmark & \cellcolor{lightblue}\pub{ICML'26} \\
        
    \bottomrule
    \end{tabular}
    }
    \label{fig:datasets}
\end{table*}

\begin{figure}[tb]
    \centering
    \includegraphics[width=1\linewidth]{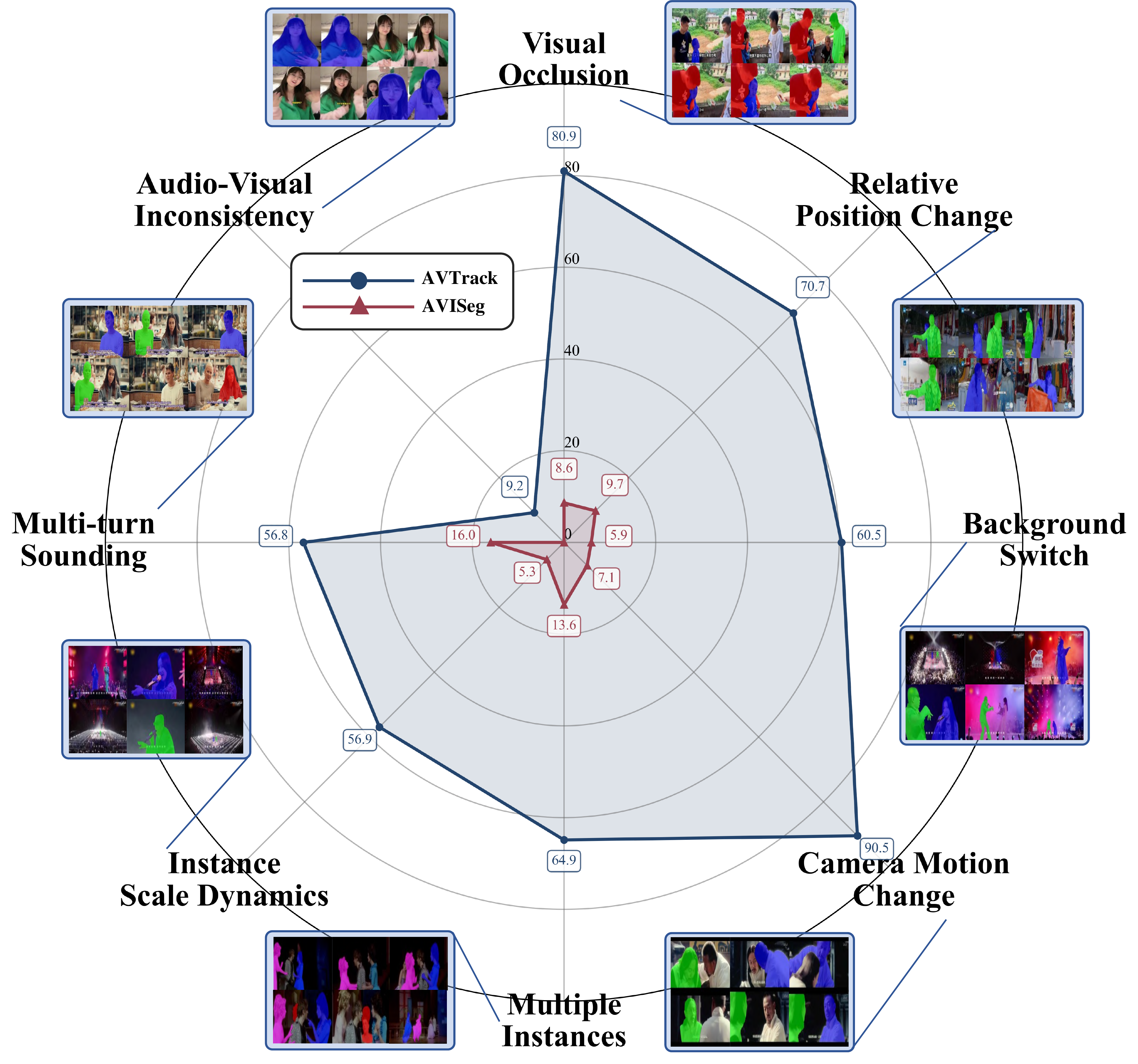}
    \caption{Comparison of AVTrack and AVISeg data distributions across different challenging conditions. Percentages indicate the proportion of each category relative to the total samples. The masked frames shown are sourced from AVTrack. \textbf{Zoom-in} for better visibility of details.}
    \label{fig:case}
    \vspace{-3mm}
\end{figure}

% AVTrack consists of 871 sounding video clips with an average duration of 54.0 seconds. Each clip is densely annotated at the pixel level, providing temporally consistent instance identities for \emph{all} human subjects throughout the video.
\cref{tab:datasets} summarizes representative benchmarks for audio-visual and visual-only detection and tracking, with a particular focus on \emph{in-the-wild} datasets that provide instance-level annotations. 
Among these, \textbf{AVTrack}, AVA-ActiveSpeaker~\cite{roth2020ava}, and YouMVOS~\cite{qi2022ovis} are explicitly designed for human-centric understanding, while differing substantially in task formulation and supervision granularity. 
AVA-ActiveSpeaker targets active speaker detection (ASD) with frame-level binary speaking labels, YouMVOS emphasizes actor-centric multi-shot tracking, and OVIS addresses general object tracking in heavily occluded scenes without audio cues. 
In contrast, AVTrack is purpose-built for \emph{human-centric AVIS} in dynamic audio-visual scenes, where complex acoustic conditions, diverse motion patterns, and rich scene composition jointly challenge spatio-temporal reasoning across modalities.

\cref{fig:case} further contrasts the distribution of challenging scenarios between AVTrack and AVISeg. Compared to AVISeg, AVTrack covers a substantially broader and more demanding spectrum of audio-visual conditions (see \cref{sec:complex_case}). In particular, AVTrack contains 9.2\% Audio-Visual Inconsistency cases and exhibits pronounced increases in visual difficulty, including Visual Occlusion (\textcolor{gray}{\small 80.9\% vs.\ 8.6\%}), instance Scale Dynamics (\textcolor{gray}{\small 56.9\% vs.\ 35.3\%}), and Background Switch (\textcolor{gray}{\small 60.5\% vs.\ 5.9\%}). Beyond appearance-level challenges, AVTrack introduces significantly more complex temporal dynamics, such as Multi-turn Sounding (\textcolor{gray}{\small 56.8\% vs.\ 16.0\%}), Relative Position Change (\textcolor{gray}{\small 70.7\% vs.\ 9.7\%}), and Camera Motion Change (\textcolor{gray}{\small 90.5\% vs.\ 7.1\%}), collectively raising the bar for modeling long-range spatio-temporal variation and cross-modal alignment.

\begin{figure}[tb]
    \centering
    \includegraphics[width=1\linewidth]{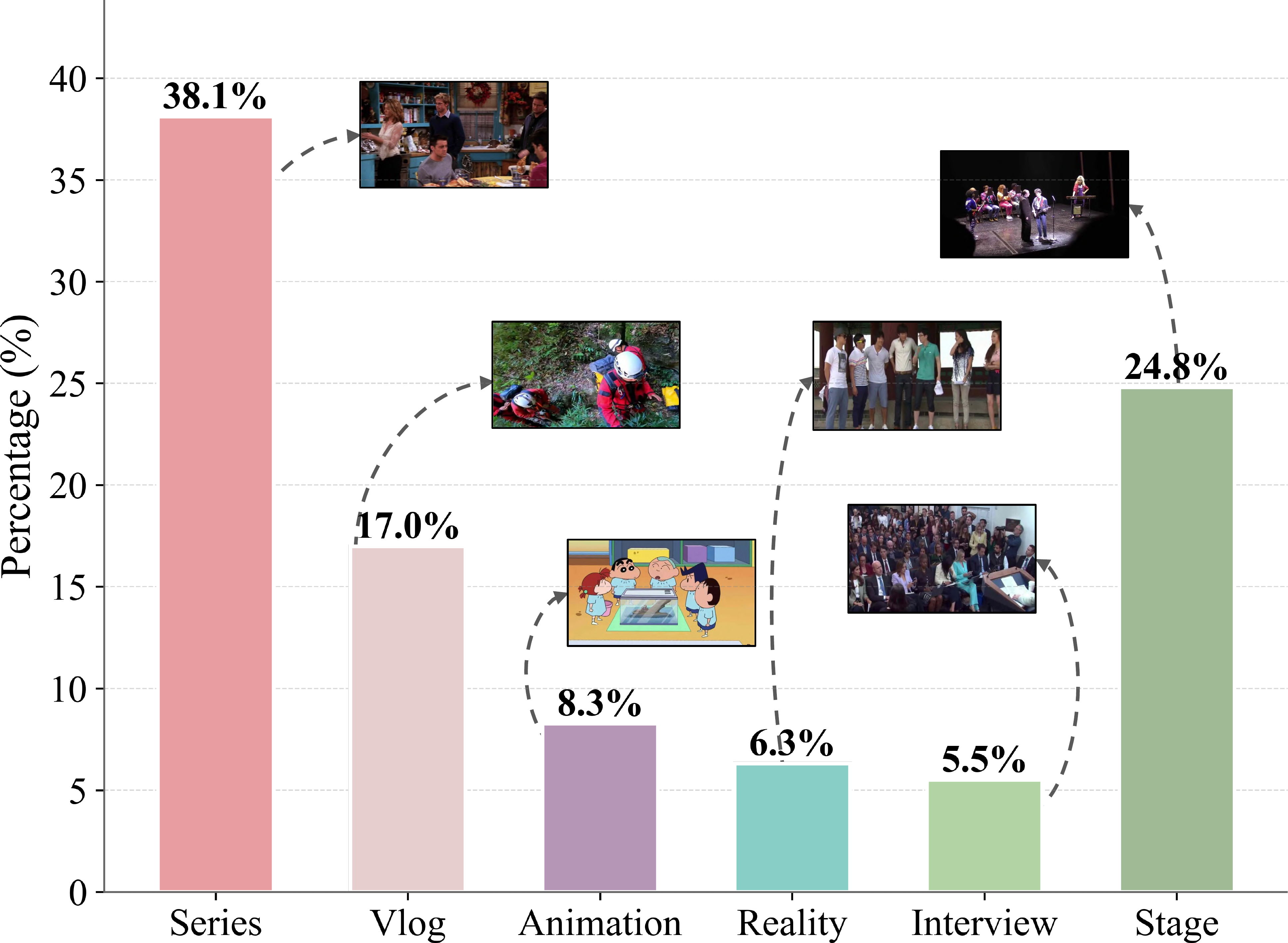}
    \caption{Video source distribution in AVTrack.}
    \label{fig:source}
\end{figure}

Another defining characteristic of AVTrack lies in the diversity of its data sources, as illustrated in \cref{fig:source}. Existing human-centric audio-visual datasets are often limited in scope: early speaker tracking benchmarks are predominantly captured in laboratory-controlled environments~\cite{lathoud2004av16,qian2019multi,qian2022audio}, while AVA-ActiveSpeaker~\cite{roth2020ava} is largely derived from Hollywood movies. In contrast, AVTrack spans a wide range of video genres, including TV series, films, vlogs, animations, reality shows, interviews, and stage performances, while deliberately incorporating complex real-world scenarios. This breadth and heterogeneity more faithfully reflect real-world audio-visual conditions, enabling comprehensive and stress-tested evaluation of model robustness, generalization, and advanced cross-modal reasoning capabilities.

% \subsection{Qualitative Dataset Comparision}
\vspace{4mm}
\section{AVTracker: A Simple Baseline}

\begin{figure*}[tb]
  \centering
  \includegraphics[width=\linewidth]{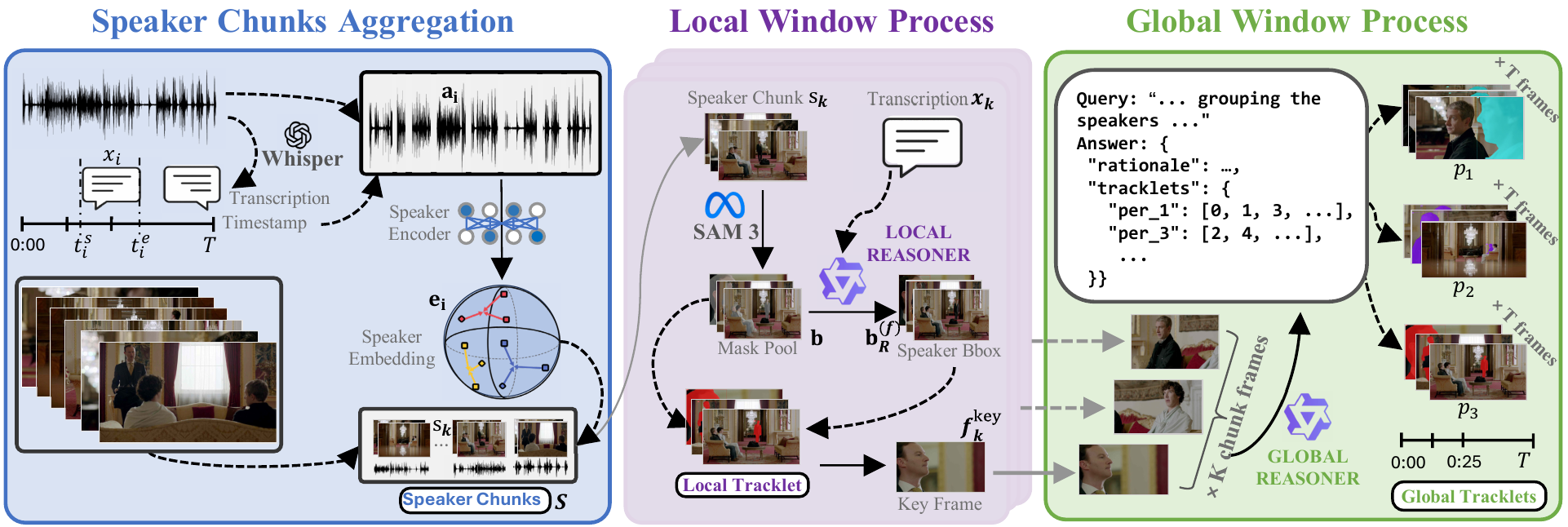}
    \caption{
        Overview of AVTracker, a three-stage framework for human-centric AVIS.
        \textbf{Stage 1 (Speaker Chunks Aggregation):}
        Speech clips are transcribed with Whisper, and timestamp-aligned transcripts together with speaker embeddings are used to group clips into speaker chunks, reducing redundancy and cost. Speech separation is ignored for clearer visualization.
        \textbf{Stage 2 (Local Window Process):}
        For each chunk, SAM3 segments candidate individuals in video frames.
        A Local Reasoner jointly conditions on speech transcripts and visual observations to associate utterances with visible persons, producing fine-grained, person-specific local tracklets.
        \textbf{Stage 3 (Global Window Process):}
        A Global Reasoner aggregates local tracklets across the entire video and resolves cross-chunk identity associations, yielding coherent instance-level speaker trajectories.
    }
  \label{fig:model_agent}
\end{figure*}

\subsection{Framework Overview}

AVTracker serves as a simple yet effective modular multi-stage baseline for human-centric AVIS in complex audio-visual scenes.
As shown in \cref{fig:model_agent}, it adopts a divide-and-conquer strategy that first constructs audio-visual alignments and local tracklets within temporally local windows, and then progressively associates tracklets of the same instance across windows to recover global speaker trajectories.
% The preceding aggregation of speaker chunks substantially reduces the number of local tracklets, leading to more efficient global window processing.

% \vspace{-10mm}
% \subsection{Implement Details}
AVTracker is implemented as a modular three-stage framework explicitly designed for extensibility, allowing new tools and functional modules to be seamlessly integrated with minimal architectural modification.
As a cascaded baseline system, AVTracker combines several advanced components to ensure strong performance across its audio-visual processing pipeline.
Specifically, we adopt Qwen3-VL \cite{bai2025qwenvl} as the visual reasoning backbone, Whisper \cite{radford2023whisper} as the speech processor, and ECAPA-TDNN \cite{desplanques2020ecapa} as the speaker embedding encoder.
To handle scenarios involving overlapping speech, MossFormer2 \cite{zhao2024mossformer2} is optionally incorporated as a speech separation module. For spatial mask generation, we employ SAM3 \cite{carion2025sam3} as the mask sampler.
Following standard practice in audio-visual learning benchmarks \cite{zhou2022avs,wang2024refavs,guo2025aviseg}, all videos are processed at a frame rate of $r = 1$ FPS. The speaker similarity threshold is set to $\tau = 0.35$.

\subsection{Speaker Chunks Aggregation}
\begin{figure}[t]
  \centering
  \includegraphics[width=\linewidth]{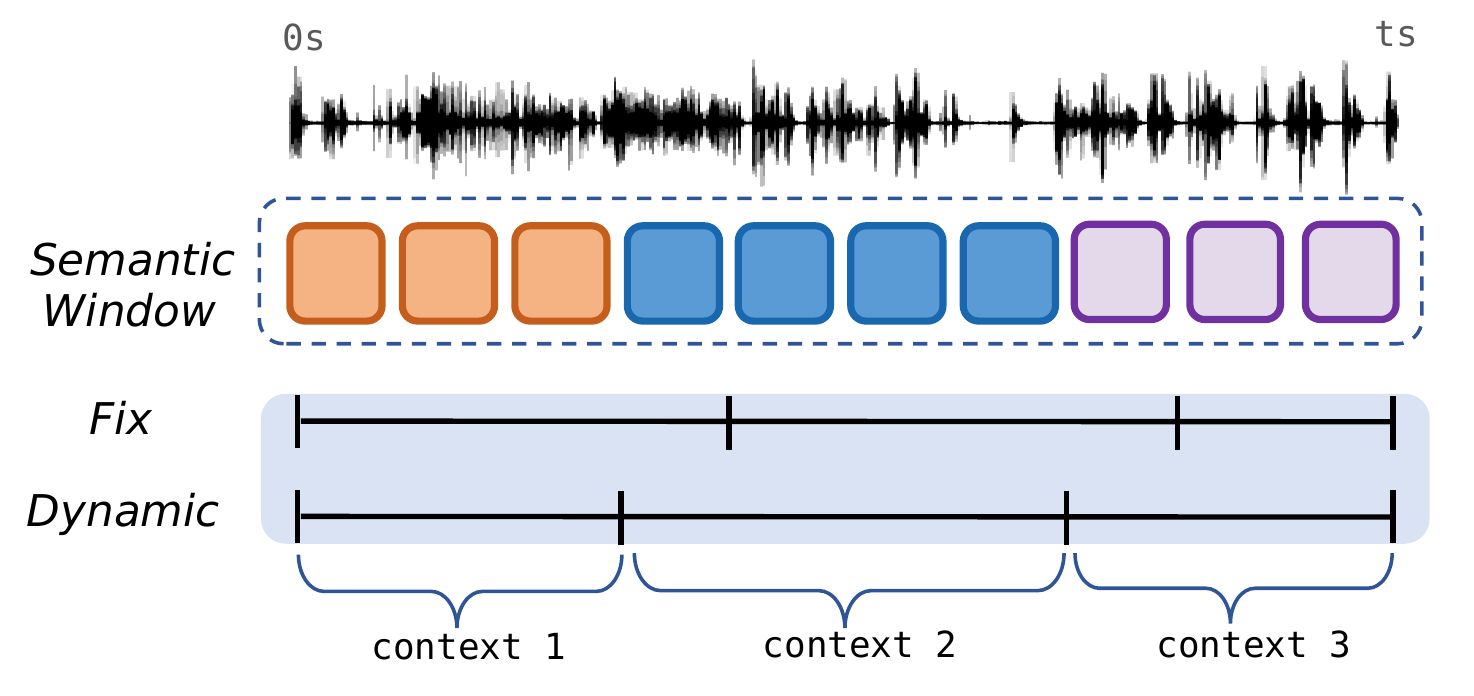}
    \caption{Comparison of windowing paradigms. Unlike fixed-size windows, dynamic windows preserve complete semantic units and contextual temporal continuity, enabling more effective and semantically coherent audio-visual correlation.}
  \label{fig:window}
\end{figure}

A natural strategy for detecting speech activity is to leverage automatic speech recognition (ASR), which transcribes audio into semantically meaningful segments and enables AVTracker to operate on dynamically sized processing windows, as illustrated in \cref{fig:window}.
Let $\mathcal{C} = \{c_i\}_{i=1}^{N}$ denote the ASR outputs including $N$ chunks, where each chunk
$c_i = (t_i^{s}, t_i^{e}, x_i)$ comprises a start time, an end time, and the transcribed text.
For each chunk, we apply a speech separation module $\mathcal{F}_{\mathrm{sep}}$ to the corresponding raw audio segment $\mathbf{a}_i$, yielding an enhanced speech signal
$\hat{\mathbf{a}}_i = \mathcal{F}_{\mathrm{sep}}(\mathbf{a}_i)$.

In practice, ASR outputs often consist of short, fragmented, and temporally adjacent segments that may originate from the same speaker.
Processing such segments independently is inefficient and increases cost, hence we aggregate adjacent segments into longer speaker chunks to reduce the number of local windows and ease the burden on the global window.
Specifically, given two consecutive chunks $c_i$ and $c_{i+1}$, we extract speaker embeddings using a pre-trained speaker encoder $\mathcal{E}$:
\vspace{-1mm}
\begin{equation}
\mathbf{e}_i = \mathcal{E}(\mathbf{\hat{a}}_i), \quad
\mathbf{e}_{i+1} = \mathcal{E}(\mathbf{\hat{a}}_{i+1}),
\end{equation}
where $\mathbf{\hat{a}}_i$ denotes the corresponding audio segment.
We then compute their cosine similarity:
\vspace{-2mm}
\begin{equation}
\text{sim}(c_i, c_{i+1}) =
\frac{\mathbf{e}_i^\top \mathbf{e}_{i+1}}
{\|\mathbf{e}_i\| \|\mathbf{e}_{i+1}\|}.
\end{equation}
% \vspace{-1mm}
If $\text{sim}(c_i, c_{i+1}) > \tau$, the two chunks are merged as
\begin{equation}
(t_i^{s}, t_i^{e}, x_i) \oplus (t_{i+1}^{s}, t_{i+1}^{e}, x_{i+1})
=
(t_i^{s}, t_{i+1}^{e}, x_i \oplus x_{i+1}),
\end{equation}
where $\oplus$ denotes temporal and textual concatenation.
Then we obtain a set of speaker chunks
$\mathcal{S} = \{s_k\}_{k=1}^{K}$.

% ------------------------------------------------------------

\subsection{Local Window Process}

The local window process associates each speaker chunk with the visible person speaking within the corresponding temporal interval.

Given a speaker chunk $s_k = (t_k^{s}, t_k^{e}, x_k)$, we convert its temporal boundaries into video frame indices.
The corresponding frame interval is computed as
\vspace{-1mm}
\begin{equation}
f_k^{s} = \left\lfloor t_k^{s} \cdot r \right\rceil, \quad
f_k^{e} = \left\lfloor t_k^{e} \cdot r \right\rceil.
\end{equation}
For the frames $\{I_f\}_{f=f_k^{s}}^{f_k^{e}}$, where $I$ is a frame, we query the Local Reasoner $\mathcal{R}^{local}$ with the speech content $x_k$ and the visual observations to localize the active speaker with a frame-wise bounding box prediction $\mathbf{b}^{(f)}_{\mathcal{R}}$:
\begin{equation}
    \mathbf{b}^{(f)}_{\mathcal{R}} = \mathcal{R}^{local}(I_f, x_k, P^{local}),
\end{equation}
where $P^{local}$ is the prompt used in local processing.
In parallel, SAM3 Video provides a set of person bounding boxes $\mathcal{B}^{(f)}_{\mathrm{SAM3}}$ and corresponding masks $\mathcal{M}^{(f)}_{\mathrm{SAM3}}$ for each frame.
We align the $\mathcal{R}^{local}$ prediction with SAM3 detections by maximizing the intersection-over-union (IoU):
\begin{equation}
\mathbf{b}^{(f)} =
\operatorname*{arg\,max}_{\mathbf{b} \in \mathcal{B}^{(f)}_{\mathrm{SAM3}}}
\ \mathrm{IoU}\!\left(\mathbf{b}^{(f)}_{\mathcal{R}}, \mathbf{b}\right),
\end{equation}
and retrieve the associated mask $\mathbf{m}^{(f)}$.
The sequence of matched masks across the chunk forms a local tracklet:
\begin{equation}
\mathcal{T}_k^{\mathrm{local}} =
\{(f, \mathbf{m}^{(f)}) \mid f_k^{s} \le f \le f_k^{e}\}.
\end{equation}
We further select a representative key frame as
\begin{equation}
f_k^{\mathrm{key}} =
\operatorname*{arg\,max}_{f \in [f_k^{s},\, f_k^{e}]}
\operatorname{Area}\!\left(\mathbf{m}^{(f)}\right),
\end{equation}
which is subsequently used for global identity association.

% ------------------------------------------------------------
\vspace{2mm}
\subsection{Global Window Process}

While local windows resolve short-term audio-visual correspondence, the same speaker may appear in multiple disjoint chunks.
The global window process aggregates local tracklets into consistent speaker identities.
Let $\mathcal{F}_{\text{key}}=\{I_{f_k^{\text{key}}}\}_{k=1}^{K}$ denote all key frames.
A Global Reasoner $\mathcal{R}^{global}$ is queried to group these frames according to person identity, producing a mapping:
\begin{equation}
\mathcal{G}: p \mapsto \{k_1, k_2, \dots\},
\end{equation}
where each identity $p$ corresponds to a set of speaker chunks.
For each identity $p$, we collect all associated local tracklets and merge them into a global trajectory:
\begin{equation}
\mathcal{T}_p =
\bigcup_{k \in \mathcal{G}(p)} \mathcal{T}_k^{\text{local}}.
\end{equation}
For frames without observations, empty masks are inserted to maintain temporal continuity.
The final output is a set of speaker trajectories
$\{\mathcal{T}_p\}$ covering the entire video.

\vspace{3mm}
\section{Experiments}
\label{sec:exp}

\subsection{Dataset Necessity and Challenges}

\begin{table*}[tb]
\centering
\footnotesize
\caption{Evaluation of VIS and AVIS methods on the AVTrack benchmark. AVIS models are trained on AVISeg, while VIS models are pretrained on YouTube-VIS and fine-tuned on AVISeg. All values are in percentage.}
\label{tab:main_exp}
\setlength\tabcolsep{10pt}
\begin{tabular}{clcccccc}
\toprule
\textbf{Task} & \textbf{Method} & \textbf{HOTA $\uparrow$} & \textbf{DetA $\uparrow$} & \textbf{AssA $\uparrow$} & \textbf{IDF1 $\uparrow$} & \textbf{MOTA $\uparrow$} & \textbf{Publication} \\
\midrule
\multirow{4}{*}{VIS}
& VITA \cite{heo2022vita} & 9.70 & 10.54 & 9.35 & 12.32 & 1.91 & \pub{NeurIPS'22} \\
& LBVQ \cite{fang2024learning} & 10.29 & 11.77 & 9.36 & 12.87 & 1.98 & \pub{TCSVT'24} \\
& CAVIS \cite{lee2025cavis} & 11.46 & 12.10 & 10.07 & 12.95 & 1.96 & \pub{ICCV'25} \\
\midrule
\multirow{4}{*}{AVIS} 
& AVISM \cite{guo2025aviseg} & 20.84 & \underline{23.22} & 19.53 & \underline{26.57} & 3.95 & \pub{CVPR'25} \\
& ACVIS \cite{seo2025learning} & 20.60 & 22.59 & 19.66 & 26.23 & \underline{4.23} & \pub{ICASSP'26} \\
& AVTrackFormer (see \cref{app:model_avtrackformer}) & \underline{21.47} & 22.51 & \underline{20.26} & 26.41 & 4.11 & -- \\
& \cellcolor{lightblue}\textbf{AVTracker} & \cellcolor{lightblue}\textbf{29.08} & \cellcolor{lightblue}\textbf{31.18} & \cellcolor{lightblue}\textbf{28.47} & \cellcolor{lightblue}\textbf{34.55} & \cellcolor{lightblue}\textbf{16.20} & \cellcolor{lightblue}\pub{ICML'26} \\
\bottomrule
\end{tabular}
\end{table*}

As shown in Table~\ref{tab:main_exp}, both VIS and AVIS methods experience pronounced performance degradation on the AVTrack benchmark. VIS-only approaches yield extremely low HOTA scores (below 12.0), indicating that visual cues alone are fundamentally insufficient for human-centric tracking in complex audio-visual scenarios.
While AVIS methods incorporate audio information and approximately double the HOTA scores of VIS baselines (remaining below 21.0), their performance is still far from satisfactory. This reveals that existing audio-visual modeling paradigms struggle to achieve robust long-term identity association under the fragmented, asynchronous, and human-centric conditions inherent to AVTrack, corresponding to the eight challenges summarized in \cref{sec:complex_case}.

In contrast, AVTracker consistently outperforms all prior methods across all evaluation metrics, achieving improvements of approximately or exceeding 8.0 points over the strongest AVIS baseline. 
These gains highlight the critical importance of robust long-range identity association in complex audio-visual scenarios.
AVTracker is designed as a modular and extensible framework, providing a strong and flexible baseline for future research on human-centric audio-visual understanding in complex scenes.

% ============== Qualitative Comparison
% \subsection{Qualitative Comparison}
% 在本section, 我们展示AVTracker相比其他模型的输出展示. 

% ============== Ablation Study
\subsection{Ablation Study}
AVTracker is designed with modularity and extensibility in mind, serving not only as a strong reference method but also as a flexible research baseline for future studies on human-centric audio-visual understanding in complex real-world scenes.
As shown in \cref{tab:ablation}, we conduct ablation studies to examine the contribution of different components in the proposed AVTracker framework. 
Specifically, we analyze the effects of (i) the model scale, (ii) the speech separation and (iii) the chunks processing.

\begin{table*}[t]
\centering
\footnotesize
\caption{
Ablation study of different model configurations on AVTrack benchmark.
244M and 809M \texttt{Param$_A$} corresponds to Whisper-small and Whisper-large-v3-turbo, respectively, while 4B and 8B \texttt{Param$_{VL}$} correspondes to Qwen3-VL-4B-Instruct and Qwen3-VL-8B-Instruct.
\textbf{Cmpr.}: whether local speech chunks are compressed to reduce computational load in the global window;
\textbf{Chunk}: whether a dynamic windowing strategy is used for chunking;
\textbf{Sepa.}: whether speech separation is applied to handle overlapping speech. 
}
\label{tab:ablation}
\setlength\tabcolsep{6pt}
\begin{tabular}{ccccccccccc}
\toprule
\textbf{Setting} & \textbf{Param$_{A}$} & \textbf{Param$_{VL}$} & \textbf{Cmpr.} & \textbf{Chunk} & \textbf{Sepa.} & \textbf{HOTA $\uparrow$} & \textbf{DetA $\uparrow$} & \textbf{AssA $\uparrow$} & \textbf{IDF1 $\uparrow$} & \textbf{MOTA $\uparrow$} \\
\midrule
% \hline

\rowcolor{lightblue}
\multicolumn{11}{c}{\textsc{AVTracker Base Setting}} \\
Base & 809M & 8B &  \cmark & \cmark & \xmarkg & 
\underline{28.85} & \textbf{31.75} & 27.39 & \underline{34.45} & \textbf{16.39} \\
\midrule

% model size
\rowcolor{lightblue}
\multicolumn{11}{c}{\textsc{Impact of Model Size}} \\

M1 & 244M & 8B &  \cmark & \cmark & \xmarkg & 
25.19 & 27.33 & 24.25 & 29.92 & 14.88 \\
M2 & 809M & 4B & \cmark & \cmark & \xmarkg & 
24.47 & 25.85 & 24.37 & 28.86 & 14.48 \\
M3 & 244M & 4B &  \cmark & \cmark & \xmarkg & 
24.01 & 25.49 & 23.69 & 28.47 & 13.52 \\
M4 & 244M & 4B/Face & \cmark & \cmark & \xmarkg & 
23.62 & 24.80 & 21.31 & 27.16 & 11.03 \\
\midrule

% separation
\rowcolor{lightblue}
\multicolumn{11}{c}{\textsc{Impact of Separation}} \\

S1 & 809M & 8B & \cmark
  & \cmark & SepFormer & 28.41  & 30.81 & \underline{27.54} & 33.65 & 15.99  \\
S2 & 809M & 8B &  \cmark & \cmark & MossFormer2 & 
\textbf{29.08} & \underline{31.18} & \textbf{28.47} & \textbf{34.55} & \underline{16.20} \\
\midrule

% chunk processing

\rowcolor{lightblue}
\multicolumn{11}{c}{\textsc{Impact of Chunk Processing}} \\
C1 & 244M & 4B & \xmarkg & \cmark & \xmarkg & 16.88 & 18.34 & 16.33 & 19.99 & 9.34 \\
C2 & 809M & 8B & \cmark
  & \xmarkg & \xmarkg & 27.45  & 29.57 & 26.64 & 32.97 & 13.49  \\
\bottomrule
\end{tabular}
\end{table*}

\myparagraph{Effect of Model Scale.}
We examine the impact of model scale in both the speech processor and the VLM reasoner.
Compared with the \texttt{Base} setting, reducing the capacity of either the speech processor (\texttt{M1}) or the VLM (\texttt{M2}) results in notable performance degradation, indicating that sufficient representational capacity in both modalities is critical for effective audio-visual reasoning.
When further reducing both the speech processor and VLM capacity in \texttt{M3}, performance drops become more pronounced, with HOTA and AssA decreasing by 4.84 and 3.70 points, respectively.
We additionally replace the VLM-based global reasoner with face detection features \cite{serengil2024benchmark} for local tracklet grouping, which leads to even larger declines on HOTA (-5.23) and AssA (-6.08).
These results show that strong audio-visual back end is essential for robust long-range identity association.

\myparagraph{Effect of Speech Separation.}
We analyze the role of speech separation for tackling potential overlapping conditions.
Compared with the \texttt{Base} setting, incorporating MossFormer \cite{zhao2024mossformer2} for speech separation (\texttt{S2}) leads to improvements in both HOTA (28.85 $\rightarrow$ 29.08) and AssA (27.39 $\rightarrow$ 28.47).
In contrast, applying with SepFormer \cite{wu2022seqformer} (\texttt{S1}) slightly degrades performance (HOTA: 28.85 $\rightarrow$ 28.41), showing that imperfect separation may introduce additional noise and temporal misalignment.
These results highlight that speech separation can be beneficial, but only when the separation quality is sufficiently reliable to support downstream audio-visual alignment.

\myparagraph{Impact of Chunk Processing.}
Removing local chunk compression (\texttt{C1}) leads to a severe performance drop compared with \texttt{M3}, with HOTA decreasing from 24.01 to 16.88, showing the importance of compact and informative local tracklets for scalable global association.
In addition, disabling dynamic windowing (\texttt{C2}) also results in a performance decline relative to the \texttt{Base} setting (HOTA: 28.85 $\rightarrow$ 27.45).

\subsection{Qualitative Comparison}

\begin{figure}[t]
  \centering
  \includegraphics[width=\linewidth]{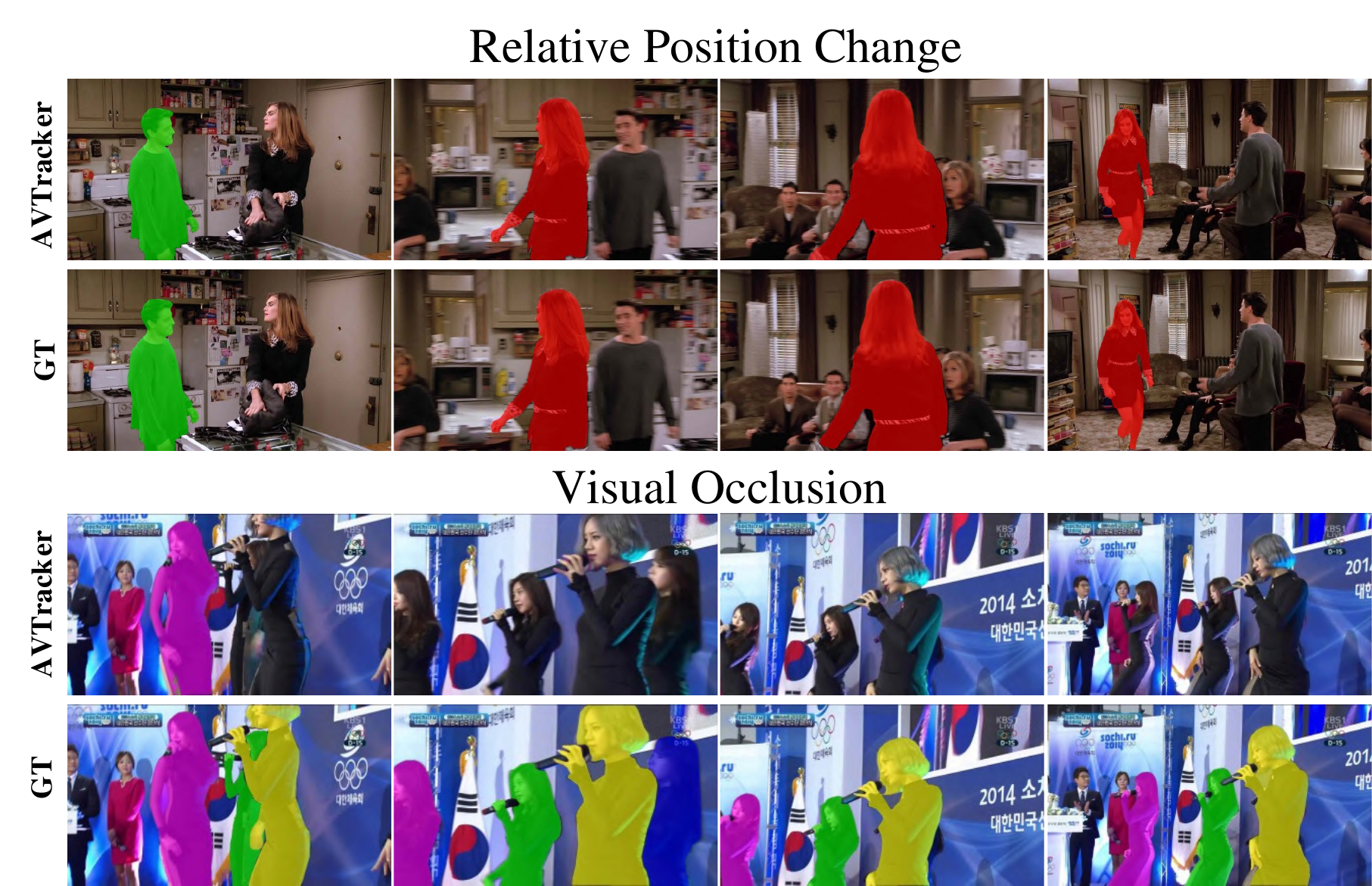}
  \caption{Qualitative comparison between AVTracker predictions and ground truth labels.}
  \label{fig:qualitative}
  \vspace{-5mm}
\end{figure}

\cref{fig:qualitative} shows a qualitative comparison between AVTracker predictions and ground-truth labels.
AVTracker tracks the active speaker well under relative position changes but struggles when multiple speakers and visual occlusions occur together. More visualizations are in \cref{app:vis}.

\section{Discussion and Future Directions}
\label{sec:future}

Based on the challenges posed by AVTrack, we outline several directions for future research in human-centric audio-visual scene understanding.

\myparagraph{Robust Audio-Visual Alignment.}
Scenarios involving multi-turn speech, visual occlusion, audio-visual inconsistencies, and multiple active speakers present extreme challenges for cross-modal alignment. A key factor in AVTracker's effectiveness is its use of textual cues as a semantic bridge between audio and visual streams. In addition, its dynamic windowing mechanism allows complete utterance-level semantic information to be captured by the visual reasoning module. Future work could explore methods that further improve alignment under such complex conditions.

\myparagraph{Audio-Visual Spatio-Temporal Intelligence.}
Dynamic changes in camera motion, relative positions, and instance scales introduce additional challenges for audio-visual crossmodal reasoning. AVTracker leverages the reasoning ability of VLM without task-specific training and carefully tuned prompts, achieving substantial performance improvements. Future research may investigate ways to enhance spatio-temporal reasoning capabilities to better handle these complex dynamics.

\myparagraph{Efficient Human-centric Data Construction.}
AVTrack is designed as a carefully constructed, test-only benchmark to provide a long-term, stable evaluation of human-centric audio-visual scene understanding. It remains an open question whether scaling up training data can help address the challenges exposed by AVTrack. Future work could focus on building efficient pipelines for human-centric data collection, potentially structured around \emph{person–time–location} triplets, and explore whether training-based approaches can bridge the observed performance gaps.

\myparagraph{Agentic Audio-Visual Reasoning.}
The challenges exposed by AVTrack, such as fragmented observations, long-term speaker identity maintenance, and ambiguous audio-visual cues, naturally motivate agentic reasoning formulations for human-centric audio-visual understanding. Rather than relying solely on one-step reasoning, future work could investigate agentic reasoning \cite{wei2026agentic} that iteratively reason over scenes through memory \cite{zhang2025survey,xu2025mem} and reflection \cite{shinn2023reflexion}. Explicit memory mechanisms may allow agents to accumulate evidence across temporal windows and recover from transient alignment errors, while reflection or self-correction \cite{pan2023automatically,gou2023critic} strategies could help revise earlier decisions when confronted with conflicting cross-modal signals. In addition, structured intermediate representations, such as dynamic scene graphs \cite{li2024scene,fei2024video} or speaker-centric relational graphs \cite{vs2023instance}, may provide a persistent abstraction for reasoning about entities and their interactions over time. 

\vspace{3mm}
\section{Conclusion}
\label{sec:conclu}
This work investigates fine-grained, human-centric audio-visual understanding and highlights the substantial gap between existing benchmarks and the complexity of real-world scenarios.
To probe the performance limits of models in realistic human-centric audio-visual environments, we introduce \textbf{AVTrack}, a challenging dataset for human-centric audio-visual instance segmentation and tracking.
AVTrack spans 6 common audio-visual data sources and covers 8 representative challenges, including visual occlusions, relative position changes, and significant camera motion.
Through comprehensive evaluations, we observe that representative AVIS methods suffer from pronounced performance degradation on AVTrack, indicating that models trained on relatively static and simplified benchmarks struggle to generalize to complex real-world settings.
To further advance this line of research, we propose a simple, effective, and extensible baseline that performs local-to-global tracklet grouping, demonstrating the feasibility of structured audio-visual reasoning and long-term tracking in challenging scenarios.
We expect AVTrack to foster research on robust, human-centric audio-visual scene understanding and to inspire the development of next-generation models capable of fine-grained spatiotemporal reasoning in real-world environments.

% Audio-Visual Reasoning, context, scene graph.
% 时空特征十分复杂

\section*{Acknowledgements} 
This work was supported by the Science and Technology Commission of Shanghai Municipality under Grant No.~25511103600 and the National Natural Science Foundation of China (NSFC) under Grant No. 62472104.

% \newpage
\section*{Impact Statement}
This work advances human-centric audio-visual tracking and scene understanding by introducing a challenging benchmark dataset and a baseline method. The proposed AVTrack aims to \textbf{support research} on robust audio-visual understanding in realistic, dynamic environments, with potential applications in video editing and human–computer interaction.
However, technologies that localize and track speaking individuals raise privacy, surveillance, and misuse concerns, especially without proper safeguards. Although AVTrack uses only publicly available videos and is \textbf{intended strictly for research}, models trained on it could be repurposed in ways that compromise privacy or enable intrusive monitoring.
We stress that \textbf{AVTrack is released solely as a research benchmark} to study current model limitations in complex audio-visual settings, not to enable real-world surveillance. We encourage future work to adopt \textbf{privacy-preserving} learning, responsible data practices, and ethical deployment. By highlighting challenges and limitations, \textit{we aim to foster more transparent, accountable, and socially responsible human-centric perception systems}.

% ``This paper presents work whose goal is to advance the field of Machine
% Learning. There are many potential societal consequences of our work, none
% which we feel must be specifically highlighted here.''

% The above statement can be used verbatim in such cases, but we encourage
% authors to think about whether there is content which does warrant further
% discussion, as this statement will be apparent if the paper is later flagged
% for ethics review.


\bibliography{example_paper}

@inproceedings{guo2025aviseg,
  title={Audio-visual instance segmentation},
  author={Guo, Ruohao and Ying, Xianghua and Chen, Yaru and Niu, Dantong and Li, Guangyao and Qu, Liao and Qi, Yanyu and Zhou, Jinxing and Xing, Bowei and Yue, Wenzhen and others},
  booktitle={Proceedings of the Computer Vision and Pattern Recognition Conference},
  pages={13550--13560},
  year={2025}
}

@inproceedings{gao2024avsegformer,
  title={Avsegformer: Audio-visual segmentation with transformer},
  author={Gao, Shengyi and Chen, Zhe and Chen, Guo and Wang, Wenhai and Lu, Tong},
  booktitle={Proceedings of the AAAI conference on artificial intelligence},
  volume={38},
  number={11},
  pages={12155--12163},
  year={2024}
}

@inproceedings{zhou2022avs,
  title={Audio--visual segmentation},
  author={Zhou, Jinxing and Wang, Jianyuan and Zhang, Jiayi and Sun, Weixuan and Zhang, Jing and Birchfield, Stan and Guo, Dan and Kong, Lingpeng and Wang, Meng and Zhong, Yiran},
  booktitle={European Conference on Computer Vision},
  pages={386--403},
  year={2022},
  organization={Springer}
}

@article{zhou2025avss,
  title={Audio-visual segmentation with semantics},
  author={Zhou, Jinxing and Shen, Xuyang and Wang, Jianyuan and Zhang, Jiayi and Sun, Weixuan and Zhang, Jing and Birchfield, Stan and Guo, Dan and Kong, Lingpeng and Wang, Meng and others},
  journal={International Journal of Computer Vision},
  volume={133},
  number={4},
  pages={1644--1664},
  year={2025},
  publisher={Springer}
}

@inproceedings{kirillov2023segment,
  title={Segment anything},
  author={Kirillov, Alexander and Mintun, Eric and Ravi, Nikhila and Mao, Hanzi and Rolland, Chloe and Gustafson, Laura and Xiao, Tete and Whitehead, Spencer and Berg, Alexander C and Lo, Wan-Yen and others},
  booktitle={Proceedings of the IEEE/CVF international conference on computer vision},
  pages={4015--4026},
  year={2023}
}

@inproceedings{wang2024refavs,
  title={Ref-avs: Refer and segment objects in audio-visual scenes},
  author={Wang, Yaoting and Sun, Peiwen and Zhou, Dongzhan and Li, Guangyao and Zhang, Honggang and Hu, Di},
  booktitle={European Conference on Computer Vision},
  pages={196--213},
  year={2024},
  organization={Springer}
}

@inproceedings{wang2024gavs,
author = {Wang, Yaoting and Liu, Weisong and Li, Guangyao and Ding, Jian and Hu, Di and Li, Xi},
title = {Prompting segmentation with sound is generalizable audio-visual source localizer},
year = {2024},
isbn = {978-1-57735-887-9},
publisher = {AAAI Press},
url = {https://doi.org/10.1609/aaai.v38i6.28378},
doi = {10.1609/aaai.v38i6.28378},
booktitle = {Proceedings of the Thirty-Eighth AAAI Conference on Artificial Intelligence and Thirty-Sixth Conference on Innovative Applications of Artificial Intelligence and Fourteenth Symposium on Educational Advances in Artificial Intelligence},
articleno = {630},
numpages = {9},
series = {AAAI'24/IAAI'24/EAAI'24}
}

@inproceedings{wang2024teso,
  title={Can Textual Semantics Mitigate Sounding Object Segmentation Preference?},
  author={Wang, Yaoting and Sun, Peiwen and Li, Yuanchao and Zhang, Honggang and Hu, Di},
  booktitle={European Conference on Computer Vision},
  pages={340--356},
  year={2024},
  organization={Springer}
}

@inproceedings{yang2024combo,
  title={Cooperation does matter: Exploring multi-order bilateral relations for audio-visual segmentation},
  author={Yang, Qi and Nie, Xing and Li, Tong and Gao, Pengfei and Guo, Ying and Zhen, Cheng and Yan, Pengfei and Xiang, Shiming},
  booktitle={Proceedings of the IEEE/CVF Conference on Computer Vision and Pattern Recognition},
  pages={27134--27143},
  year={2024}
}

@inproceedings{yang2019video,
  title={Video instance segmentation},
  author={Yang, Linjie and Fan, Yuchen and Xu, Ning},
  booktitle={Proceedings of the IEEE/CVF international conference on computer vision},
  pages={5188--5197},
  year={2019}
}

@article{qi2022ovis,
  title={Occluded video instance segmentation: A benchmark},
  author={Qi, Jiyang and Gao, Yan and Hu, Yao and Wang, Xinggang and Liu, Xiaoyu and Bai, Xiang and Belongie, Serge and Yuille, Alan and Torr, Philip HS and Bai, Song},
  journal={International Journal of Computer Vision},
  volume={130},
  number={8},
  pages={2022--2039},
  year={2022},
  publisher={Springer}
}

@inproceedings{athar2020stem,
  title={Stem-seg: Spatio-temporal embeddings for instance segmentation in videos},
  author={Athar, Ali and Mahadevan, Sabarinath and Osep, Aljosa and Leal-Taix{\'e}, Laura and Leibe, Bastian},
  booktitle={European conference on computer vision},
  pages={158--177},
  year={2020},
  organization={Springer}
}

@inproceedings{cao2020sipmask,
  title={Sipmask: Spatial information preservation for fast image and video instance segmentation},
  author={Cao, Jiale and Anwer, Rao Muhammad and Cholakkal, Hisham and Khan, Fahad Shahbaz and Pang, Yanwei and Shao, Ling},
  booktitle={European Conference on Computer Vision},
  pages={1--18},
  year={2020},
  organization={Springer}
}

@inproceedings{liu2021sg,
  title={Sg-net: Spatial granularity network for one-stage video instance segmentation},
  author={Liu, Dongfang and Cui, Yiming and Tan, Wenbo and Chen, Yingjie},
  booktitle={Proceedings of the IEEE/CVF conference on computer vision and pattern recognition},
  pages={9816--9825},
  year={2021}
}

@inproceedings{yang2021crossover,
  title={Crossover learning for fast online video instance segmentation},
  author={Yang, Shusheng and Fang, Yuxin and Wang, Xinggang and Li, Yu and Fang, Chen and Shan, Ying and Feng, Bin and Liu, Wenyu},
  booktitle={proceedings of the IEEE/CVF international conference on computer vision},
  pages={8043--8052},
  year={2021}
}

@inproceedings{he2017mask,
  title={Mask r-cnn},
  author={He, Kaiming and Gkioxari, Georgia and Doll{\'a}r, Piotr and Girshick, Ross},
  booktitle={Proceedings of the IEEE international conference on computer vision},
  pages={2961--2969},
  year={2017}
}

@inproceedings{guo2021sotr,
  title={Sotr: Segmenting objects with transformers},
  author={Guo, Ruohao and Niu, Dantong and Qu, Liao and Li, Zhenbo},
  booktitle={Proceedings of the IEEE/CVF international conference on computer vision},
  pages={7157--7166},
  year={2021}
}

@inproceedings{wang2021end,
  title={End-to-end video instance segmentation with transformers},
  author={Wang, Yuqing and Xu, Zhaoliang and Wang, Xinlong and Shen, Chunhua and Cheng, Baoshan and Shen, Hao and Xia, Huaxia},
  booktitle={Proceedings of the IEEE/CVF conference on computer vision and pattern recognition},
  pages={8741--8750},
  year={2021}
}

@inproceedings{wu2022seqformer,
  title={Seqformer: Sequential transformer for video instance segmentation},
  author={Wu, Junfeng and Jiang, Yi and Bai, Song and Zhang, Wenqing and Bai, Xiang},
  booktitle={European Conference on Computer Vision},
  pages={553--569},
  year={2022},
  organization={Springer}
}

@inproceedings{cheng2022masked,
  title={Masked-attention mask transformer for universal image segmentation},
  author={Cheng, Bowen and Misra, Ishan and Schwing, Alexander G and Kirillov, Alexander and Girdhar, Rohit},
  booktitle={Proceedings of the IEEE/CVF conference on computer vision and pattern recognition},
  pages={1290--1299},
  year={2022}
}

@article{hwang2021video,
  title={Video instance segmentation using inter-frame communication transformers},
  author={Hwang, Sukjun and Heo, Miran and Oh, Seoung Wug and Kim, Seon Joo},
  journal={Advances in Neural Information Processing Systems},
  volume={34},
  pages={13352--13363},
  year={2021}
}

@article{heo2022vita,
  title={Vita: Video instance segmentation via object token association},
  author={Heo, Miran and Hwang, Sukjun and Oh, Seoung Wug and Lee, Joon-Young and Kim, Seon Joo},
  journal={Advances in neural information processing systems},
  volume={35},
  pages={23109--23120},
  year={2022}
}

@inproceedings{ding2023mose,
  title={MOSE: A new dataset for video object segmentation in complex scenes},
  author={Ding, Henghui and Liu, Chang and He, Shuting and Jiang, Xudong and Torr, Philip HS and Bai, Song},
  booktitle={Proceedings of the IEEE/CVF international conference on computer vision},
  pages={20224--20234},
  year={2023}
}

@inproceedings{zha2025implicit,
  title={Implicit Counterfactual Learning for Audio-Visual Segmentation},
  author={Zha, Mingfeng and Li, Tianyu and Wang, Guoqing and Wang, Peng and Wu, Yangyang and Yang, Yang and Shen, Heng Tao},
  booktitle={Proceedings of the IEEE/CVF International Conference on Computer Vision},
  pages={22349--22360},
  year={2025}
}

@inproceedings{senocak2018learning,
  title={Learning to localize sound source in visual scenes},
  author={Senocak, Arda and Oh, Tae-Hyun and Kim, Junsik and Yang, Ming-Hsuan and Kweon, In So},
  booktitle={Proceedings of the IEEE conference on computer vision and pattern recognition},
  pages={4358--4366},
  year={2018}
}

@article{zhao2023audio,
  title={Audio-visual speaker tracking: Progress, challenges, and future directions},
  author={Zhao, Jinzheng and Xu, Yong and Qian, Xinyuan and Berghi, Davide and Wu, Peipei and Cui, Meng and Sun, Jianyuan and Jackson, Philip JB and Wang, Wenwu},
  journal={arXiv preprint arXiv:2310.14778},
  year={2023}
}

@inproceedings{li2022multi,
  title={Multi-modal perception attention network with self-supervised learning for audio-visual speaker tracking},
  author={Li, Yidi and Liu, Hong and Tang, Hao},
  booktitle={Proceedings of the AAAI conference on artificial intelligence},
  year={2022}
}

@inproceedings{li2025multi,
  title={Multi-Stage Multimodal Distillation for Audio-Visual Speaker Tracking},
  author={Li, Yidi and Zhao, Wenkai and Wang, Zeyu and Xu, Zhenhuan and Ren, Bin and Sebe, Nicu},
  booktitle={ICASSP 2025-2025 IEEE International Conference on Acoustics, Speech and Signal Processing (ICASSP)},
  pages={1--5},
  year={2025},
  organization={IEEE}
}

@inproceedings{lee2025cavis,
  title={CAVIS: Context-Aware Video Instance Segmentation},
  author={Lee, Seunghun and Seo, Jiwan and Han, Kiljoon and Choi, Minwoo and Im, Sunghoon},
  booktitle={Proceedings of the IEEE/CVF International Conference on Computer Vision},
  pages={4507--4517},
  year={2025}
}

@inproceedings{lathoud2004av16,
  title={AV16. 3: An audio-visual corpus for speaker localization and tracking},
  author={Lathoud, Guillaume and Odobez, Jean-Marc and Gatica-Perez, Daniel},
  booktitle={International Workshop on Machine Learning for Multimodal Interaction},
  pages={182--195},
  year={2004},
  organization={Springer}
}

@article{qian2019multi,
  title={Multi-speaker tracking from an audio--visual sensing device},
  author={Qian, Xinyuan and Brutti, Alessio and Lanz, Oswald and Omologo, Maurizio and Cavallaro, Andrea},
  journal={IEEE Transactions on Multimedia},
  volume={21},
  number={10},
  pages={2576--2588},
  year={2019},
  publisher={IEEE}
}

@article{qian2022audio,
  title={Audio-visual cross-attention network for robotic speaker tracking},
  author={Qian, Xinyuan and Wang, Zhengdong and Wang, Jiadong and Guan, Guohui and Li, Haizhou},
  journal={IEEE/ACM Transactions on Audio, Speech, and Language Processing},
  volume={31},
  pages={550--562},
  year={2022},
  publisher={IEEE}
}

@inproceedings{chen2021localizing,
  title={Localizing visual sounds the hard way},
  author={Chen, Honglie and Xie, Weidi and Afouras, Triantafyllos and Nagrani, Arsha and Vedaldi, Andrea and Zisserman, Andrew},
  booktitle={Proceedings of the IEEE/CVF conference on computer vision and pattern recognition},
  pages={16867--16876},
  year={2021}
}

@inproceedings{yang2019vis,
  title={Video instance segmentation},
  author={Yang, Linjie and Fan, Yuchen and Xu, Ning},
  booktitle={Proceedings of the IEEE/CVF international conference on computer vision},
  pages={5188--5197},
  year={2019}
}

@article{ren2024grounded,
  title={Grounded sam: Assembling open-world models for diverse visual tasks},
  author={Ren, Tianhe and Liu, Shilong and Zeng, Ailing and Lin, Jing and Li, Kunchang and Cao, He and Chen, Jiayu and Huang, Xinyu and Chen, Yukang and Yan, Feng and others},
  journal={arXiv preprint arXiv:2401.14159},
  year={2024}
}

@inproceedings{roth2020ava,
  title={Ava active speaker: An audio-visual dataset for active speaker detection},
  author={Roth, Joseph and Chaudhuri, Sourish and Klejch, Ondrej and Marvin, Radhika and Gallagher, Andrew and Kaver, Liat and Ramaswamy, Sharadh and Stopczynski, Arkadiusz and Schmid, Cordelia and Xi, Zhonghua and others},
  booktitle={ICASSP 2020-2020 IEEE international conference on acoustics, speech and signal processing (ICASSP)},
  pages={4492--4496},
  year={2020},
  organization={IEEE}
}

@article{fang2024learning,
  title={Learning better video query with sam for video instance segmentation},
  author={Fang, Hao and Zhang, Tong and Zhou, Xiaofei and Zhang, Xinxin},
  journal={IEEE Transactions on Circuits and Systems for Video Technology},
  year={2024},
  publisher={IEEE}
}

@article{seo2025learning,
  title={Learning What To Hear: Boosting Sound-Source Association For Robust Audiovisual Instance Segmentation},
  author={Seo, Jinbae and Kwon, Hyeongjun and Kim, Kwonyoung and Lee, Jiyoung and Sohn, Kwanghoon},
  journal={arXiv preprint arXiv:2509.22740},
  year={2025}
}

@article{luiten2021hota,
  title={Hota: A higher order metric for evaluating multi-object tracking},
  author={Luiten, Jonathon and Osep, Aljosa and Dendorfer, Patrick and Torr, Philip and Geiger, Andreas and Leal-Taix{\'e}, Laura and Leibe, Bastian},
  journal={International journal of computer vision},
  volume={129},
  number={2},
  pages={548--578},
  year={2021},
  publisher={Springer}
}

@inproceedings{gavrilyuk2018actor,
  title={Actor and action video segmentation from a sentence},
  author={Gavrilyuk, Kirill and Ghodrati, Amir and Li, Zhenyang and Snoek, Cees GM},
  booktitle={Proceedings of the IEEE conference on computer vision and pattern recognition},
  pages={5958--5966},
  year={2018}
}

@inproceedings{Giancola2018soccernet,
   title={SoccerNet: A Scalable Dataset for Action Spotting in Soccer Videos},
   url={http://dx.doi.org/10.1109/CVPRW.2018.00223},
   DOI={10.1109/cvprw.2018.00223},
   booktitle={2018 IEEE/CVF Conference on Computer Vision and Pattern Recognition Workshops (CVPRW)},
   publisher={IEEE},
   author={Giancola, Silvio and Amine, Mohieddine and Dghaily, Tarek and Ghanem, Bernard},
   year={2018},
   month=jun, pages={1792–179210} }

@inproceedings{deliege2021soccernet,
  title={Soccernet-v2: A dataset and benchmarks for holistic understanding of broadcast soccer videos},
  author={Deliege, Adrien and Cioppa, Anthony and Giancola, Silvio and Seikavandi, Meisam J and Dueholm, Jacob V and Nasrollahi, Kamal and Ghanem, Bernard and Moeslund, Thomas B and Van Droogenbroeck, Marc},
  booktitle={Proceedings of the IEEE/CVF conference on computer vision and pattern recognition},
  pages={4508--4519},
  year={2021}
}

@inproceedings{Cioppa2022SoccerNetTracking,
   title={SoccerNet-Tracking: Multiple Object Tracking Dataset and Benchmark in Soccer Videos},
   url={http://dx.doi.org/10.1109/cvprw56347.2022.00393},
   DOI={10.1109/cvprw56347.2022.00393},
   booktitle={2022 IEEE/CVF Conference on Computer Vision and Pattern Recognition Workshops (CVPRW)},
   publisher={IEEE},
   author={Cioppa, Anthony and Giancola, Silvio and Deliege, Adrien and Kang, Le and Zhou, Xin and Cheng, Zhiyu and Ghanem, Bernard and Van Droogenbroeck, Marc},
   year={2022},
   month=jun, pages={3490–3501} }

@inproceedings{wei2022youmvos,
  title={Youmvos: an actor-centric multi-shot video object segmentation dataset},
  author={Wei, Donglai and Kharbanda, Siddhant and Arora, Sarthak and Roy, Roshan and Jain, Nishant and Palrecha, Akash and Shah, Tanav and Mathur, Shray and Mathur, Ritik and Kemkar, Abhijay and others},
  booktitle={Proceedings of the IEEE/CVF Conference on Computer Vision and Pattern Recognition},
  pages={21044--21053},
  year={2022}
}

@article{lu2025llava,
  title={LLaVA-ReID: Selective Multi-image Questioner for Interactive Person Re-Identification},
  author={Lu, Yiding and Yang, Mouxing and Peng, Dezhong and Hu, Peng and Lin, Yijie and Peng, Xi},
  journal={arXiv preprint arXiv:2504.10174},
  year={2025}
}

@article{hu2024empowering,
  title={Empowering visible-infrared person re-identification with large foundation models},
  author={Hu, Zhangyi and Yang, Bin and Ye, Mang},
  journal={Advances in Neural Information Processing Systems},
  volume={37},
  pages={117363--117387},
  year={2024}
}

@inproceedings{tan2024harnessing,
  title={Harnessing the power of mllms for transferable text-to-image person reid},
  author={Tan, Wentan and Ding, Changxing and Jiang, Jiayu and Wang, Fei and Zhan, Yibing and Tao, Dapeng},
  booktitle={Proceedings of the IEEE/CVF Conference on Computer Vision and Pattern Recognition},
  pages={17127--17137},
  year={2024}
}

@article{yang2024mllmreid,
  title={MLLMReID: multimodal large language model-based person re-identification},
  author={Yang, Shan and Zhang, Yongfei},
  journal={arXiv preprint arXiv:2401.13201},
  year={2024}
}

@article{bai2025qwenvl,
  title={Qwen2. 5-vl technical report},
  author={Bai, Shuai and Chen, Keqin and Liu, Xuejing and Wang, Jialin and Ge, Wenbin and Song, Sibo and Dang, Kai and Wang, Peng and Wang, Shijie and Tang, Jun and others},
  journal={arXiv preprint arXiv:2502.13923},
  year={2025}
}

@inproceedings{radford2023whisper,
  title={Robust speech recognition via large-scale weak supervision},
  author={Radford, Alec and Kim, Jong Wook and Xu, Tao and Brockman, Greg and McLeavey, Christine and Sutskever, Ilya},
  booktitle={International conference on machine learning},
  pages={28492--28518},
  year={2023},
  organization={PMLR}
}

@article{desplanques2020ecapa,
  title={Ecapa-tdnn: Emphasized channel attention, propagation and aggregation in tdnn based speaker verification},
  author={Desplanques, Brecht and Thienpondt, Jenthe and Demuynck, Kris},
  journal={arXiv preprint arXiv:2005.07143},
  year={2020}
}

@inproceedings{zhao2024mossformer2,
  title={Mossformer2: Combining transformer and rnn-free recurrent network for enhanced time-domain monaural speech separation},
  author={Zhao, Shengkui and Ma, Yukun and Ni, Chongjia and Zhang, Chong and Wang, Hao and Nguyen, Trung Hieu and Zhou, Kun and Yip, Jia Qi and Ng, Dianwen and Ma, Bin},
  booktitle={ICASSP 2024-2024 IEEE International Conference on Acoustics, Speech and Signal Processing (ICASSP)},
  pages={10356--10360},
  year={2024},
  organization={IEEE}
}

@article{carion2025sam3,
  title={Sam 3: Segment anything with concepts},
  author={Carion, Nicolas and Gustafson, Laura and Hu, Yuan-Ting and Debnath, Shoubhik and Hu, Ronghang and Suris, Didac and Ryali, Chaitanya and Alwala, Kalyan Vasudev and Khedr, Haitham and Huang, Andrew and others},
  journal={arXiv preprint arXiv:2511.16719},
  year={2025}
}

@inproceedings{he2021transreid,
  title={Transreid: Transformer-based object re-identification},
  author={He, Shuting and Luo, Hao and Wang, Pichao and Wang, Fan and Li, Hao and Jiang, Wei},
  booktitle={Proceedings of the IEEE/CVF international conference on computer vision},
  pages={15013--15022},
  year={2021}
}

@inproceedings{yi2014deep,
  title={Deep metric learning for person re-identification},
  author={Yi, Dong and Lei, Zhen and Liao, Shengcai and Li, Stan Z},
  booktitle={2014 22nd international conference on pattern recognition},
  pages={34--39},
  year={2014},
  organization={IEEE}
}

@inproceedings{zhao2017deeply,
  title={Deeply-learned part-aligned representations for person re-identification},
  author={Zhao, Liming and Li, Xi and Zhuang, Yueting and Wang, Jingdong},
  booktitle={Proceedings of the IEEE international conference on computer vision},
  pages={3219--3228},
  year={2017}
}

@article{li2021person,
  title={Person re-identification with part prediction alignment},
  author={Li, Zhiyong and Lv, Jingyi and Chen, Ying and Yuan, Jin},
  journal={Computer Vision and Image Understanding},
  volume={205},
  pages={103172},
  year={2021},
  publisher={Elsevier}
}

@article{serengil2024benchmark,
  title={A benchmark of facial recognition pipelines and co-usability performances of modules},
  author={Serengil, Sefik and {\"O}zp{\i}nar, Alper},
  journal={Bili{\c{s}}im Teknolojileri Dergisi},
  volume={17},
  number={2},
  pages={95--107},
  year={2024},
  publisher={Gazi University}
}

@article{zhang2025survey,
  title={A survey on the memory mechanism of large language model-based agents},
  author={Zhang, Zeyu and Dai, Quanyu and Bo, Xiaohe and Ma, Chen and Li, Rui and Chen, Xu and Zhu, Jieming and Dong, Zhenhua and Wen, Ji-Rong},
  journal={ACM Transactions on Information Systems},
  volume={43},
  number={6},
  pages={1--47},
  year={2025},
  publisher={ACM New York, NY}
}

@article{xu2025mem,
  title={A-mem: Agentic memory for llm agents},
  author={Xu, Wujiang and Liang, Zujie and Mei, Kai and Gao, Hang and Tan, Juntao and Zhang, Yongfeng},
  journal={arXiv preprint arXiv:2502.12110},
  year={2025}
}

@article{shinn2023reflexion,
  title={Reflexion: Language agents with verbal reinforcement learning},
  author={Shinn, Noah and Cassano, Federico and Gopinath, Ashwin and Narasimhan, Karthik and Yao, Shunyu},
  journal={Advances in Neural Information Processing Systems},
  volume={36},
  pages={8634--8652},
  year={2023}
}

@article{wei2026agentic,
  title={Agentic Reasoning for Large Language Models},
  author={Wei, Tianxin and Li, Ting-Wei and Liu, Zhining and Ning, Xuying and Yang, Ze and Zou, Jiaru and Zeng, Zhichen and Qiu, Ruizhong and Lin, Xiao and Fu, Dongqi and others},
  journal={arXiv preprint arXiv:2601.12538},
  year={2026}
}

@article{pan2023automatically,
  title={Automatically correcting large language models: Surveying the landscape of diverse self-correction strategies},
  author={Pan, Liangming and Saxon, Michael and Xu, Wenda and Nathani, Deepak and Wang, Xinyi and Wang, William Yang},
  journal={arXiv preprint arXiv:2308.03188},
  year={2023}
}

@article{gou2023critic,
  title={Critic: Large language models can self-correct with tool-interactive critiquing},
  author={Gou, Zhibin and Shao, Zhihong and Gong, Yeyun and Shen, Yelong and Yang, Yujiu and Duan, Nan and Chen, Weizhu},
  journal={arXiv preprint arXiv:2305.11738},
  year={2023}
}

@article{li2024scene,
author = {Li, Hongsheng and Zhu, Guangming and Zhang, Liang and Jiang, Youliang and Dang, Yixuan and Hou, Haoran and Shen, Peiyi and Zhao, Xia and Shah, Syed Afaq Ali and Bennamoun, Mohammed},
title = {Scene Graph Generation: A comprehensive survey},
year = {2024},
issue_date = {Jan 2024},
publisher = {Elsevier Science Publishers B. V.},
address = {NLD},
volume = {566},
number = {C},
issn = {0925-2312},
url = {https://doi.org/10.1016/j.neucom.2023.127052},
doi = {10.1016/j.neucom.2023.127052},
journal = {Neurocomput.},
month = jan,
numpages = {25},
}

@article{fei2024video,
  title={Video-of-thought: Step-by-step video reasoning from perception to cognition},
  author={Fei, Hao and Wu, Shengqiong and Ji, Wei and Zhang, Hanwang and Zhang, Meishan and Lee, Mong-Li and Hsu, Wynne},
  journal={arXiv preprint arXiv:2501.03230},
  year={2024}
}

@inproceedings{vs2023instance,
  title={Instance relation graph guided source-free domain adaptive object detection},
  author={VS, Vibashan and Oza, Poojan and Patel, Vishal M},
  booktitle={Proceedings of the IEEE/CVF conference on computer vision and pattern recognition},
  pages={3520--3530},
  year={2023}
}

@inproceedings{carion2020end,
  title={End-to-end object detection with transformers},
  author={Carion, Nicolas and Massa, Francisco and Synnaeve, Gabriel and Usunier, Nicolas and Kirillov, Alexander and Zagoruyko, Sergey},
  booktitle={European conference on computer vision},
  pages={213--229},
  year={2020},
  organization={Springer}
}

@inproceedings{liu2021swin,
  title={Swin transformer: Hierarchical vision transformer using shifted windows},
  author={Liu, Ze and Lin, Yutong and Cao, Yue and Hu, Han and Wei, Yixuan and Zhang, Zheng and Lin, Stephen and Guo, Baining},
  booktitle={Proceedings of the IEEE/CVF international conference on computer vision},
  pages={10012--10022},
  year={2021}
}
\bibliographystyle{icml2026}

%%%%%%%%%%%%%%%%%%%%%%%%%%%%%%%%%%%%%%%%%%%%%%%%%%%%%%%%%%%%%%%%%%%%%%%%%%%%%%%
%%%%%%%%%%%%%%%%%%%%%%%%%%%%%%%%%%%%%%%%%%%%%%%%%%%%%%%%%%%%%%%%%%%%%%%%%%%%%%%
% APPENDIX
%%%%%%%%%%%%%%%%%%%%%%%%%%%%%%%%%%%%%%%%%%%%%%%%%%%%%%%%%%%%%%%%%%%%%%%%%%%%%%%
%%%%%%%%%%%%%%%%%%%%%%%%%%%%%%%%%%%%%%%%%%%%%%%%%%%%%%%%%%%%%%%%%%%%%%%%%%%%%%%
\newpage 
% \null
% \newpage 
\appendix
% \onecolumn

% ======================== Related Works =====================
\newpage
\null
\newpage 

\section{Additional Related Works}
\subsection{Video Instance Segmentation}
Video instance segmentation (VIS) \cite{yang2019video} is the task of simultaneously detecting, segmenting, and tracking instances across frames in a video.
Early VIS methods~\cite{yang2019video,cao2020sipmask,athar2020stem,liu2021sg,yang2021crossover} mainly evolve from CNN-based image instance segmentation frameworks~\cite{he2017mask,guo2021sotr}, incorporating temporal cues via optical flow, feature propagation, or tracking modules.
For example, MaskTrack R-CNN~\cite{yang2019video} extends Mask R-CNN~\cite{he2017mask} with an association branch for temporal linking, while SGNet~\cite{liu2021sg} adopts an anchor-free centerness-based design for robust correspondence.
Transformer-based architectures then bring a paradigm shift toward end-to-end VIS. VisTR~\cite{wang2021end} first extends DETR~\cite{carion2020end} into a query-based formulation that directly predicts instance mask sequences, removing explicit association. Subsequent models like SeqFormer~\cite{wu2022seqformer} and Mask2Former-VIS~\cite{cheng2022masked} enhance query interaction and multi-scale aggregation, achieving stronger segmentation and tracking. Token-centric frameworks such as IFC~\cite{hwang2021video} and VITA~\cite{heo2022vita} further improve efficiency and scalability by compressing spatio-temporal features into compact instance-aware tokens, enabling efficient long-range reasoning across videos.
CAVIS \cite{lee2025cavis} introduces a context-aware approach to VIS, enhancing object tracking and segmentation accuracy by incorporating surrounding contextual information across frames.

\subsection{Human-centric Object Tracking}
Several human-centric video understanding datasets have been proposed to study people, actions, and interactions in complex real-world videos. SoccerNet~\cite{Giancola2018soccernet} introduces large-scale broadcast soccer videos for action spotting, focusing on temporally localized human actions and game events in long, untrimmed sequences. SoccerNet-v2~\cite{deliege2021soccernet} extends this line of work toward a more holistic understanding of soccer broadcasts by incorporating richer annotations and multiple complementary tasks, encouraging joint modeling of human actions, camera dynamics, and high-level semantics. Building further on human-centered analysis, SoccerNet-Tracking~\cite{Cioppa2022SoccerNetTracking} shifts the focus to multiple object tracking, emphasizing consistent identity modeling of players, referees, and other key actors over time in crowded and dynamic scenes. Outside the sports domain, YouMVOS~\cite{wei2022youmvos} addresses actor-centric video object segmentation in unconstrained videos, highlighting the challenge of maintaining coherent actor-centric instance segmentation across multiple shots and scene changes. Together, these works reflect a progression from sparse action recognition toward comprehensive, human-centric understanding of long and complex videos.

% - https://arxiv.org/abs/1804.04527 SoccerNet: A Scalable Dataset for Action Spotting in Soccer Videos 
% - https://arxiv.org/abs/2011.13367 SoccerNet-v2: A Dataset and Benchmarks for Holistic Understanding of Broadcast Soccer Videos 
% - https://arxiv.org/abs/2204.06918 SoccerNet-Tracking: Multiple Object Tracking Dataset and Benchmark in Soccer Videos 
% - YouMVOS: An Actor-centric Multi-shot Video Object Segmentation Dataset
\subsection{Person Re-identification (Re-ID)}

Person Re-ID seeks to match images of the same individual across views or video shots, facing challenges from illumination, viewpoint, pose variations, occlusions, and background clutter. Traditional approaches focus on learning discriminative visual features, using attention mechanisms~\cite{he2021transreid} to reduce background noise, part-based or locally aligned representations~\cite{zhao2017deeply,li2021person} to handle pose misalignment, and metric learning~\cite{yi2014deep} to structure identity embeddings.
The advent of VLMs has inspired integrating multimodal semantic information into ReID. VLMs pretrained on image–text pairs can capture high-level attributes such as clothing, accessories, or demographic cues, complementing visual features for open-vocabulary retrieval or text-based search.
LLaVa-ReID~\cite{lu2025llava} iteratively queries fine-grained attributes to refine ambiguous descriptions. TVI-LFM~\cite{hu2024empowering} leverages VLM- and LLM-generated text for joint modality alignment, improving infrared features and cross-modal consistency. MLLM4Text-ReID~\cite{tan2024harnessing} automatically generates diverse textual templates and enhances cross-dataset transfer without target-domain fine-tuning. MLLMReID~\cite{yang2024mllmreid} adapts visual encoders through a unified instruction strategy and synchronized multi-task training.
This line of work suggests that leveraging high-level semantic reasoning improves robustness and generalization in challenging ReID settings, paralleling our emphasis on human-centric tracking.

\section{Data Collection and Annotation Pipeline}
\label{sec:appendix_annotation}

\begin{figure*}[tb]
  \centering
  \includegraphics[width=\linewidth]{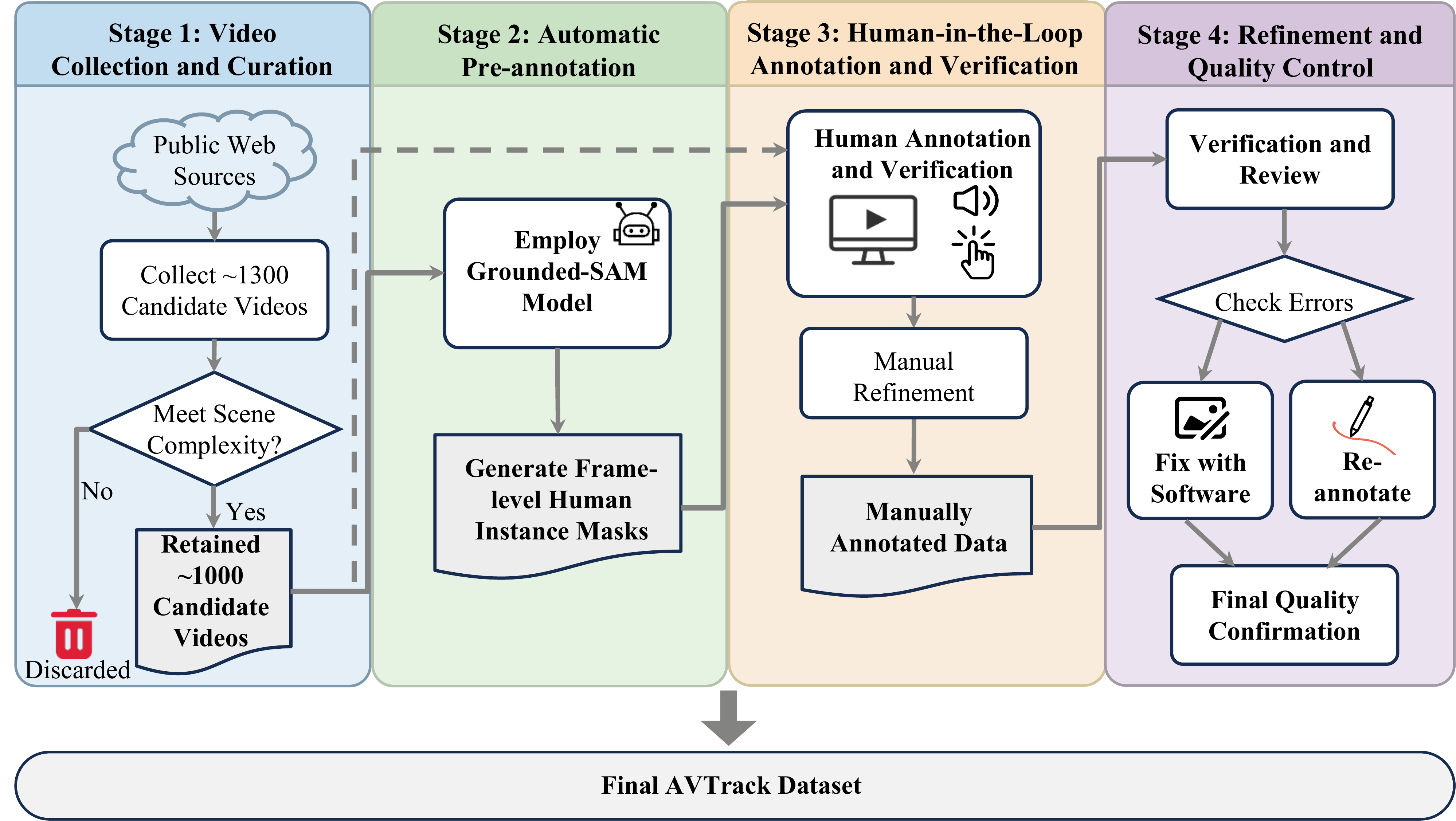}
  \caption{AVTrack dataset construction pipeline.}
  \label{fig:pipeline}
\end{figure*}

As shown in \cref{fig:pipeline}, AVTrack is built through a carefully designed pipeline consisting of four key stages: 

\myparagraph{Video Collection and Curation.}
We first collect approximately 1,300 candidate videos from publicly available web sources, targeting real-world human-centric scenarios such as interviews and performances. 
These videos are subjected to an initial screening stage that removes samples failing to meet minimum requirements on the scene complexity mentioned above. 
After filtering, roughly 1,000 videos are retained as candidates for annotation. This curation step ensures that the dataset consistently reflects non-trivial audio-visual interactions beyond simple and static environments.

\myparagraph{Automatic Pre-annotation.}
To bootstrap instance-level annotations at scale, we employ Grounded-SAM \cite{ren2024grounded} to generate frame-level human instance masks for each candidate video. This automatic stage provides coarse yet comprehensive spatial coverage and significantly reduces the manual annotation burden, while preserving the flexibility required for subsequent human refinement.

\myparagraph{Human-in-the-Loop Annotation and Verification.}
To ensure annotation consistency and correctness, all annotators undergo structured training on a held-out subset of the data. Each annotator is required to complete an initial annotation pass, which is reviewed by the project team before full-scale labeling begins. We further develop a dedicated web-based annotation tool that enables annotators to track individual instances across frames and manually add or correct masks missed by the automatic process. During annotation, synchronized video, audio, and full-frame visual context are provided, and annotators are instructed to label \emph{sounding human instances} on a per-frame basis.

\myparagraph{Refinement and Quality Control.}
Following manual annotation, each video undergoes a refinement stage to correct residual errors. Minor artifacts, such as fragmented regions or small holes, are fixed using graphic editing software, while severely inaccurate or missing instances are re-annotated using LabelMe \footnote{LabelMe: https://github.com/wkentaro/labelme}. This multi-level verification process ensures spatial accuracy, temporal consistency, and reliable audio-visual correspondence across frames.

\myparagraph{Final Dataset.}  
After rigorous quality control, the final AVTrack benchmark comprises \textbf{871 high-quality video clips} encompassing 3,120 annotated instances. Over a period of nearly three months, 15 professional annotators contributed to data collection, annotation, and verification. 
This meticulous pipeline resulted in a test-only dataset whose annotation fidelity and scene complexity surpass those of existing human-centric audio-visual benchmarks, providing a reliable and challenging evaluation platform for advancing robust audio-visual perception and long-term identity association in complex real-world scenarios.

\section{Evaluation Metrics}
\label{app:evalmet}
We evaluate human-centric AVIS using the TrackEval toolkit~\cite{luiten2021hota}, which assesses detection accuracy, temporal association, and identity consistency. 
These capabilities are critical in human-centric scenarios involving frequent speaker switches, occlusions, and long-range interactions. 
Following prior work~\cite{cheng2022masked,heo2022vita,guo2025aviseg}, we adopt five standard evaluation metrics: HOTA (Higher-Order Tracking Accuracy), DetA (Detection Accuracy), AssA (Association Accuracy), IDF1 (Identity F1 Score), and MOTA (Multiple Object Tracking Accuracy).
All metrics are computed using the official API to ensure reproducibility and fair comparison, reported per video, and averaged across the dataset. 

\myparagraph{HOTA.}
Higher Order Tracking Accuracy (HOTA) jointly measures detection and association performance by averaging scores over different matching thresholds. Formally, HOTA is defined as
\begin{equation}
\mathrm{HOTA} = \frac{1}{|\mathcal{A}|} \sum_{\alpha \in \mathcal{A}} 
\sqrt{\mathrm{DetA}_\alpha \cdot \mathrm{AssA}_\alpha},
\end{equation}
where $\alpha$ denotes the matching threshold, and $\mathcal{A}$ is the set of thresholds used for evaluation. By explicitly balancing detection and association terms, HOTA provides a holistic assessment of tracking quality.

\myparagraph{Detection Accuracy (DetA).}
Detection Accuracy evaluates frame-level instance detection correctness and is defined as
\begin{equation}
\mathrm{DetA} = \frac{\mathrm{TP}}{\mathrm{TP} + \mathrm{FN} + \mathrm{FP}},
\end{equation}
where $\mathrm{TP}$, $\mathrm{FN}$, and $\mathrm{FP}$ denote the numbers of true positives, false negatives, and false positives, respectively, based on the matching criterion used for evaluation. 
In the context of AVIS, DetA measures whether sounding human instances are correctly localized at the frame level.

\myparagraph{Association Accuracy (AssA).}
Association Accuracy evaluates the correctness of temporal associations between matched detections across frames. It is defined as
\begin{equation}
\mathrm{AssA} = \frac{1}{|\mathcal{C}|} \sum_{c \in \mathcal{C}} 
\frac{|\mathrm{TPA}(c)|}{|\mathrm{TPA}(c)| + |\mathrm{FNA}(c)| + |\mathrm{FPA}(c)|},
\end{equation}
where $\mathcal{C}$ denotes the set of true positive detection matches obtained under the adopted matching criterion, and $\mathrm{TPA}(c)$, $\mathrm{FNA}(c)$, and $\mathrm{FPA}(c)$ represent the numbers of true positive, false negative, and false positive associations induced by the matched detection $c$, respectively. 
AssA quantifies local association correctness and penalizes inconsistent temporal linkages, reflecting a model’s ability to maintain coherent instance associations over time.

\myparagraph{IDF1.}
IDF1 evaluates global identity preservation and is defined as
\begin{equation}
\mathrm{IDF1} = \frac{2 \cdot \mathrm{IDTP}}
{2 \cdot \mathrm{IDTP} + \mathrm{IDFP} + \mathrm{IDFN}},
\end{equation}
where $\mathrm{IDTP}$, $\mathrm{IDFP}$, and $\mathrm{IDFN}$ denote the numbers of identity true positives, false positives, and false negatives, respectively. IDF1 is sensitive to long-term identity fragmentation and is particularly relevant for challenging dynamic human-centric scenes.

\myparagraph{MOTA.}
Multi-Object Tracking Accuracy (MOTA) aggregates detection and identity errors into a single metric:
\begin{equation}
\mathrm{MOTA} = 1 - 
\frac{\mathrm{FN} + \mathrm{FP} + \mathrm{IDSW}}{\mathrm{GT}},
\end{equation}
where $\mathrm{IDSW}$ is the number of identity switches and $\mathrm{GT}$ denotes the total number of ground-truth instances.

\begin{figure*}[tb]
  \centering
  \includegraphics[width=\linewidth]{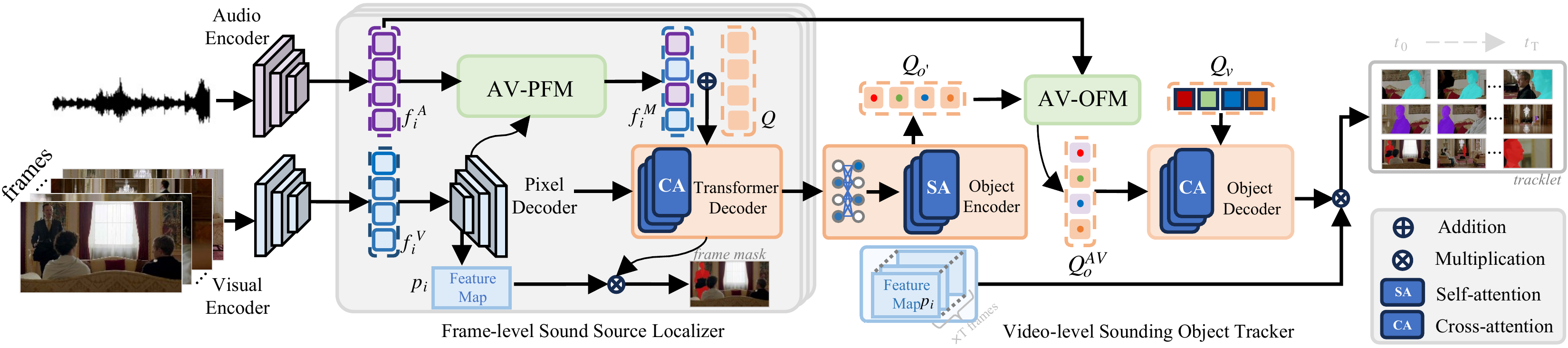}
  \caption{An overview of AVTrackFormer: similar to AVISM, it first performs pixel-level cross-modal fusion through AV-PFM. However, AVTrackFormer enables bidirectional interaction between object tokens and audio features in AV-OFM, rather than a unidirectional one.}
  \label{fig:trackformer}
\end{figure*}

\begin{figure}[tb]
  \centering
  \includegraphics[width=1\linewidth]{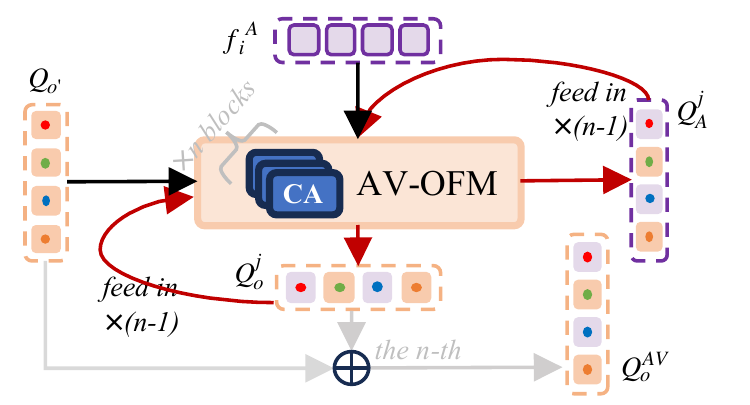}
  \caption{Audio-Visual Object-level Fusion Module (AV-OFM). AV-OFM is designed to model audio–visual correlations in an interleaved and bidirectional manner.}
  \label{fig:ofm}
\end{figure}

\section{AVTrackFormer: An End-to-end Baseline}
\label{app:model_avtrackformer}

Upon observing AVISM's performance on AVTrack, we conducted a preliminary experiment by making minor modifications to AVISM, yielding a new variant named AVTrackFormer. Unlike AVTracker, AVTrackFormer is an end-to-end model trained in the same manner as AVISM \cite{guo2025aviseg} but features enhanced modeling of audio-visual interaction during the video-level sounding object tracking.

\subsection{Audio-Visual Representation}
\label{sec:av_representation}
Given an input video with temporally aligned visual and audio streams, we segment it into $T$ consecutive, non-overlapping temporal snippets $\{(v_i, a_i)\}_{i=1}^{T}$, each spanning one second.  
For each visual snippet $v_i$, a visual backbone extracts multi-scale feature maps $f^{V}_{i,k} \in \mathbb{R}^{H_k \times W_k \times D_k}$, where $k$ indexes the stages of the backbone. The aggregated visual representation across all snippets is denoted as $F_V = \{ f^{V}_{i} \}_{i=1}^{T}$.  
For each audio snippet $a_i$, we compute a log-mel spectrogram, which is then encoded into a fixed-dimensional embedding $f^{A}_{i} \in \mathbb{R}^{D}$ using a pre-trained audio encoder. The resulting audio representation for the video is denoted as $F_A = \{ f^{A}_{i} \}_{i=1}^{T}$. The parameters of the audio encoder are kept frozen during training.

\subsection{Frame-Level Sound Source Localizer}
\label{sec:frame_level_localizer}

We perform frame-wise localization of sounding objects by explicitly modeling spatial audio-visual correspondence. Given the visual features $f^{V}_{i}$, a pixel decoder generates enhanced visual representations $\hat{f}^{V}_{i}$ as well as per-pixel embeddings $p_i \in \mathbb{R}^{H \times W \times C}$.  
To integrate audio information, the Audio-Visual Pixel-level Fusion Module (AV-PFM) applies cross-attention between $\hat{f}^{V}_{i}$ and the corresponding audio embedding $f^{A}_{i}$, producing audio-conditioned visual features $f^{M}_{i} \in \mathbb{R}^{C}$.  
Following the set prediction paradigm~\cite{carion2020end,guo2025aviseg}, we introduce $N_f$ learnable frame queries $Q \in \mathbb{R}^{N_f \times C}$, each conditioned on the multimodal feature $f^{M}_{i}$, resulting in frame queries $Q_f \in \mathbb{R}^{N_f \times C}$.  
These queries are further refined through a Transformer decoder. Subsequently, each query is projected into the pixel embedding space via a dot product with $p_i$ to predict both the class label and segmentation mask corresponding to each sounding object.

\subsection{Video-Level Sounding Object Tracker}
\label{sec:video_level_tracker}

This module associates frame-level predictions across an entire video. In line with AVISM and to mitigate computational overhead in long or high-resolution videos, the tracker operates on frame queries rather than dense pixel features. Concretely, a linear layer projects the $T \times N_f$ frame queries from all frames into object tokens $Q_o$.  
Following VITA~\cite{heo2022vita}, these object tokens are processed by an object encoder using windowed self-attention~\cite{liu2021swin}, producing encoded tokens ${Q}_{o'}$.  
Audio cues are incorporated into the temporal modeling via the Audio-Visual Object-level Fusion Module (AV-OFM) applied to ${Q}_{o'}$, resulting in audio-conditioned object tokens $Q^{AV}_{o}$. 
To extract object-centric representations from all tokens, we initialize a fixed set of learnable video queries $Q_v \in \mathbb{R}^{N_v \times C}$, where $N_v$ denotes the number of video queries. A transformer decoder serves as the object decoder, which takes $Q_o^{AV}$ as input and aggregates their semantic information into the video queries. The decoder output are then dot-multiplied with $p_i$ to get the final mask logits.

\myparagraph{AV-OFM:} 
In contrast to AVISM, where cross-modal interaction is limited to a unidirectional enhancement, specifically, audio features $f_i^A$ only attend to object tokens $Q_{o'}$ to update themselves before being fused back, AVTrackFormer introduces the enhanced AV-OFM to enable bidirectional interactions.  
As illustrated in \cref{fig:ofm}, our AV-OFM treats object tokens $Q_{o'}$ and audio features $f_i^A$ as co-evolving states. We denote the intermediate states at the $j$-th block as $Q_o^j$ and $Q_A^j$, with initial states $Q_o^0 = Q_{o'}$ and $Q_A^0 = f_i^A$. For each iteration $j \in \{1, \dots, n-1\}$, the module recursively updates both modalities:  
\begin{equation}
(Q_o^j, Q_A^j) = \text{AV-OFM}(Q_o^{j-1}, Q_A^{j-1}).
\end{equation}
Within this iterative process, the cross-attention mechanism (\texttt{CA}) facilitates reciprocal information exchange, where object tokens are conditioned on audio context and audio features are simultaneously refined using object-level visual cues. After the final $n$-th block, the output $Q_o^n$ is combined with the original input $Q_{o'}$ via a residual connection to yield the final fused representation:  
\begin{equation}
Q_o^{AV} = Q_{o'} \oplus Q_o^n.
\end{equation}

% two heads are employed for final prediction, each consisting of two fully-connected layers. Specifically, the class head outputs class probabilities p∈RK+1p \in \mathbb{R}^{K+1} for each video query, where the additional class accounts for “no sounding object” alongside the KK dataset classes. Meanwhile, the mask head takes the object queries and computes a dot product with pip_i, yielding the final mask logits.

% ===
% Within each local window, the module computes cross-attention between Qo′{Q}_{o'} and the corresponding audio features {fAi}\{ f^{A}_{i} \}. After stacking NN attention layers, the output tokens are added to Qo′{Q}_{o'}, yielding audio-conditioned object tokens QAVoQ^{AV}_{o}.

% Two prediction heads are applied to each video query: a classification head producing probabilities p∈RK+1p \in \mathbb{R}^{K+1}, where the extra class represents ``no sounding object'', and a mask head whose outputs are dot-multiplied with pip_i to generate the final segmentation masks.

\subsection{Impact of Different Visual Backbone}
\begin{table}[tb]
\centering
\caption{Impact of visual backbone on AVTrack. All models are evaluated with backbone pretrained on ImageNet and Coco dataset.}
\label{tab:backbone_impact}
\resizebox{\columnwidth}{!}{
\begin{tabular}{lcccccc}
\toprule
\textbf{Backbone} & \textbf{Model} & \textbf{HOTA} & \textbf{DetA} & \textbf{AssA} & \textbf{IDF1} & \textbf{MOTA} \\
\midrule
\multirow{3}{*}{R-50} 
& AVISM & 16.56  & 17.39 & 16.48 & 21.59 & 1.89 \\
& ACVIS & 16.94  & 18.02  & 16.62  & 22.55 & 2.18 \\
&  \cellcolor{lightblue}AVTrackFormer & \cellcolor{lightblue}17.50 & \cellcolor{lightblue}18.91 & \cellcolor{lightblue}16.97 & \cellcolor{lightblue}22.78 & \cellcolor{lightblue}2.25 \\
\midrule
\multirow{3}{*}{Swin-L} 
& AVISM & \underline{20.84} & \textbf{23.22} & 19.53 & \textbf{26.57} & 3.95 \\
& ACVIS & 20.60 & \underline{22.59} & \underline{19.66} & 26.23 & \textbf{4.23} \\
& \cellcolor{lightblue}AVTrackFormer & \cellcolor{lightblue}\textbf{21.47} & \cellcolor{lightblue}22.51 & \cellcolor{lightblue}\textbf{20.26} & \cellcolor{lightblue}\underline{26.41} & \cellcolor{lightblue}\underline{4.11} \\
\bottomrule
\end{tabular}%
}
\end{table}

\cref{tab:backbone_impact} analyzes the impact of different visual backbones on AVTrack and representative AVIS models. Overall, replacing the ResNet-50 backbone with the stronger Swin-Large consistently leads to substantial performance improvements across all evaluated metrics and models, highlighting the critical role of high-capacity visual representations in complex audio-visual tracking scenarios.

Under the ResNet-50 backbone, all methods exhibit relatively limited performance, with AVTrackFormer achieving the best overall results among the three models, Indicating that more effective audio-visual interaction design can yield consistent gains in both detection quality and identity preservation.
When adopting the Swin-Large backbone, the performance gap between methods becomes more nuanced. All models benefit notably from stronger visual features, with absolute improvements of approximately 3-4 points in HOTA compared to ResNet-50. AVTrackFormer attains the highest HOTA and AssA scores, suggesting superior association capability when richer visual context is available, while AVISM and ACVIS achieve competitive results in DetA and MOTA, respectively. These results suggest that stronger backbones primarily enhance detection robustness, whereas improvements in association accuracy depend more heavily on the tracking and fusion design.

\section{Qualitative Comparison}
\label{app:vis}

\begin{strip}
  \centering
  \includegraphics[width=\linewidth]{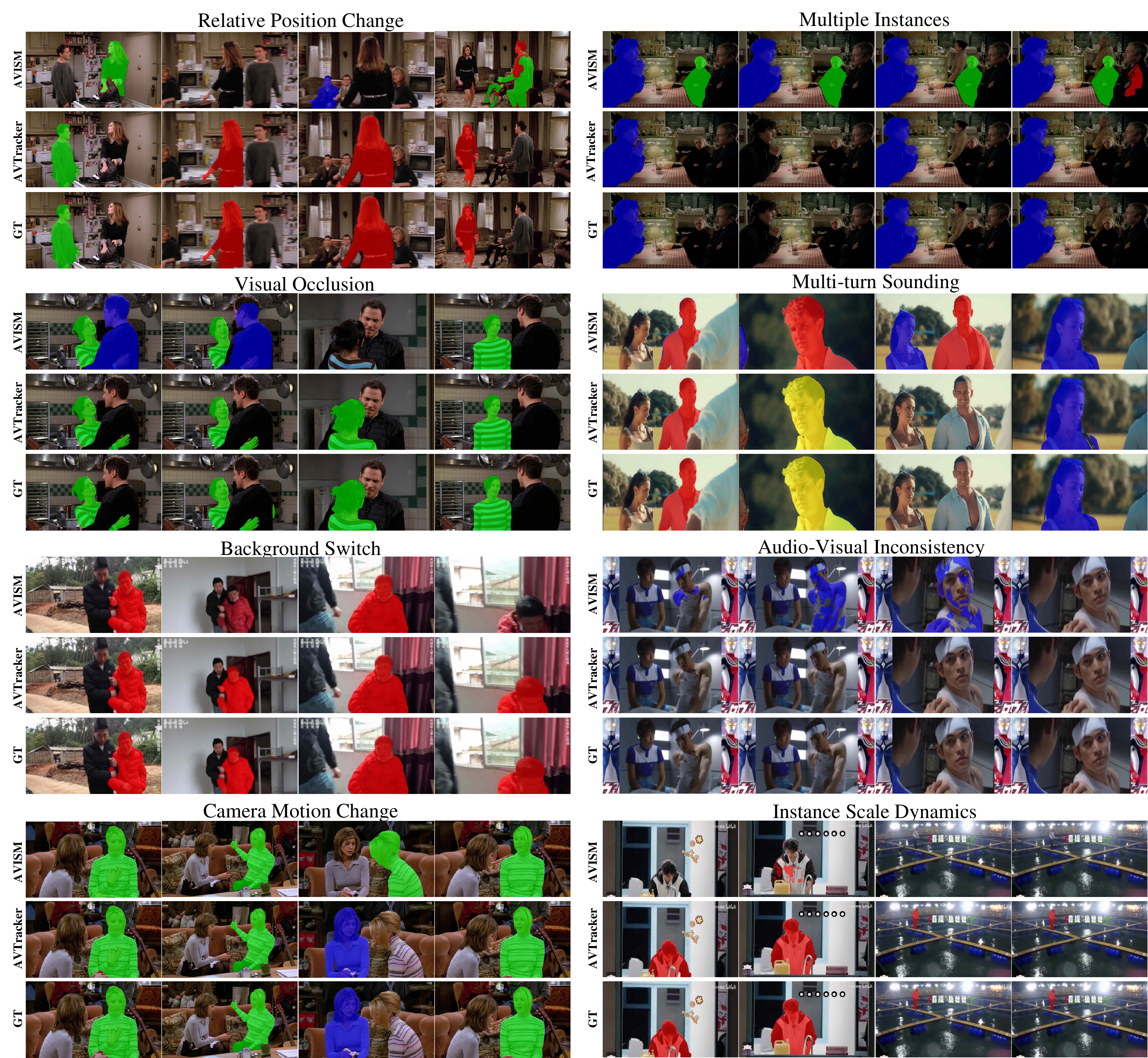}
  \captionof{figure}{Qualitative comparison between AVTracker and AVISM.}
  \label{fig:full_vis}
\end{strip}

As shown in \cref{fig:full_vis}, in this section, we provides qualitative comparison across AVTracker and AVISM under different challenging cases.

% 在 Relative Position Change 场景下，AVISM 因跨帧身份关联能力相对有限，偶尔会错误追踪到沉默的实例, 更值得注意的是, 在首帧和尾帧中, 因为相对位置的变化, 将不同的说话人设定为同一个人；在 Camera Motion Change 场景中，AVISM 对相机运动干扰的鲁棒性有限，可能出现音频-视觉关联错误；
% 在 Multi-turn Sounding 场景下，AVISM 对多轮发声建模能力有限，可能将同一实例 ID 错误绑定到另一说话人，导致 ID shift；
% 而 AVTracker受益于模块化的先分割再匹配, 以及local-to-global tracklet grouping机制, 可以较好的实现speaker identity assignment.

% 在 Visual Occlusion 场景下，AVISM 在遮挡条件下对实例完整性的建模较弱，比如在第一帧和第二帧中, 错误的把女性的手分配给男性；在 Background Switch 场景下，AVISM 对背景变化较敏感，容易导致speaker detection不稳定. 
% 在 Instance Scale Dynamics 场景下，AVISM 对小尺度实例的响应较弱，远距离小尺寸目标的检测与跟踪效果有限.
% 在 Multiple Instances 场景下，AVISM 难以在外观相似或空间邻近的多人中建立精准的音频-视觉对应关系，可能出现实例混淆；
% 而 AVTracker 借助强大的预训练分割基础模型, 对上述情况的处理更为鲁棒。

% 在 Audio-Visual Inconsistency 场景下，AVISM 的音频-视觉联合理解能力相对弱，可能将语音来源错误归因于画面中视觉显著但实际静默的对象；而 AVTracker 通过dynamic windowing处理语义信息, 将speech语义与视觉线索进行关联，能够准确处理Audio-Visual Inconsistency.
In the \textbf{Relative Position Change} scenario, changes in relative positioning between speakers in the first and last frames may lead AVISM to mistakenly treat different individuals as the same. In the \textbf{Camera Motion Change} scenario, AVISM has limited robustness to camera motion, potentially resulting in incorrect audio-visual associations. In the \textbf{Multi-turn Sounding} scenario, AVISM struggles with modeling multiple rounds of speech, often erroneously assigning the same instance ID to different speakers, which causes identity shifts. In contrast, AVTracker's modular ``segment-then-match'' design, coupled with a local-to-global tracklet grouping mechanism, supports accurate speaker identity assignment, handling these challenges more effectively.

In the \textbf{Visual Occlusion} scenario, AVISM demonstrates weak modeling of instance integrity when occlusion occurs. For example, it may incorrectly assign a woman's hand to a man in the first and second frames. In the \textbf{Background Switch} scenario, AVISM is sensitive to background changes, which leads to unstable speaker detection. In the \textbf{Instance Scale Dynamics} scenario, AVISM has limited performance for small-scale instances, leading to suboptimal detection and tracking of distant or small targets. In the \textbf{Multiple Instances} scenario, AVISM struggles to establish precise audio-visual correspondences between individuals with similar appearances or those located close to each other, resulting in instance confusion. In contrast, AVTracker effectively handles these issues, leveraging a robust pre-trained segmentation backbone to enhance performance.

In the \textbf{Audio-Visual Inconsistency} scenario, AVISM has weak joint audio-visual understanding, occasionally attributing speech to visually salient but silent objects. On the other hand, AVTracker employs dynamic windowing to integrate semantic information, effectively linking speech semantics with visual cues and accurately resolving audio-visual inconsistencies.

% \subsection{Failure Case Analysis}
% 在所提的challenging case下的分析? 如何写, 感觉所提的case很大程度都是依靠llm的分析, 难不成说是llm的分析是瓶颈. 

\section{Implementation Details}
\label{sec:impl_details}
 
\subsection{Model Configuration}
AVTracker is composed entirely of off-the-shelf pretrained models, which makes the pipeline easy to reproduce and to upgrade as stronger components become available. The specific variants used in all reported experiments are listed in \cref{tab:model_config}.
 
\begin{table}[h]
\centering
\caption{Model configuration of AVTracker.}
\label{tab:model_config}
\resizebox{\columnwidth}{!}{
\begin{tabular}{lll}
\toprule
\textbf{Component} & \textbf{Model} & \textbf{Purpose} \\
\midrule
VLM               & Qwen3-VL-8B-Instruct      & Local/Global Reasoner \\
ASR               & Whisper-large-v3-turbo    & Speech transcription \\
Speaker encoder   & ECAPA-TDNN (SpeechBrain)  & Embedding + aggregation \\
Speech separator  & MossFormer2 (ModelScope)  & Overlapping speech \\
Mask generator    & SAM3-Video (Meta)         & Instance mask generation \\
\bottomrule
\end{tabular}
}
\end{table}
 
\subsection{Hyperparameters and Hardware}
The speaker similarity threshold is set to $\tau = 0.35$, the input frame rate is $r = 1$~FPS, and the IoU matching threshold for tracklet association is $0.3$. All experiments are run on a single 48\,GB NVIDIA A6000 GPU.
\cref{tab:tau_sensitivity} sweeps the speaker similarity threshold $\tau$. The plateau region $\tau \in [0.30, 0.40]$ varies by only $0.56$ HOTA, with our default $\tau = 0.35$ sitting at its center. The value was fixed early in development using a quick scan on five videos and never re-tuned, indicating that AVTracker is robust to this hyperparameter.
 
\begin{table}[h]
\centering
\caption{Sensitivity to the speaker similarity threshold $\tau$.}
\label{tab:tau_sensitivity}
\begin{tabular}{cccc}
\toprule
\textbf{$\tau$} & \textbf{HOTA} & \textbf{DetA} & \textbf{AssA} \\
\midrule
0.20 & 26.55 & 28.09 & 26.71 \\
0.25 & 28.09 & 29.21 & 28.70 \\
0.30 & 28.53 & 30.63 & 28.14 \\
\rowcolor{lightgray}
0.35 & 29.08 & 31.18 & 28.47 \\
0.40 & 29.09 & 29.96 & 28.66 \\
0.45 & 26.46 & 27.63 & 26.41 \\
0.50 & 27.34 & 28.22 & 27.65 \\
\bottomrule
\end{tabular}
\end{table}

\section{Local and Global Reasoner}
The two VLM reasoners are driven by carefully designed prompts. We summarize their core specifications below; the complete prompt texts are released with the code.
 
\textbf{Local Reasoner.} Given video frames and the corresponding speech transcript, the reasoner decides whether the speech originates from a visible person in the frame, conditioning on facial features, lip motion, hairstyle, and clothing. Candidate persons are marked with red bounding boxes. When a match exists, the reasoner emits the speaker's bounding box for each frame; otherwise it emits $[0,0,0,0]$. The output is a JSON object with two fields: \texttt{rationale} (free-form reasoning) and \texttt{boxes} (per-frame coordinates).
 
\textbf{Global Reasoner.} Given persons across frames, the reasoner determines which instances belong to the same individual. It groups frames by identity and assigns each person a unique label (\texttt{person\_0}, \texttt{person\_1}, $\ldots$). The output is a JSON object with two fields: \texttt{rationale} and \texttt{persons}, where \texttt{persons} maps each \texttt{person\_id} to its frame indices.

\section{Computational Complexity and Efficiency}
\label{sec:complexity}
 
% \paragraph{End-to-end vs.\ pipeline cost.}
AVTracker trades inference speed for accuracy by leveraging a VLM. \cref{tab:runtime_complexity} compares it against end-to-end baselines, and \cref{tab:flops_breakdown} attributes its cost to individual components.
 
\begin{table}[h]
\centering
\caption{Parameter count, per-frame FLOPs, and throughput.}
\label{tab:runtime_complexity}
\small
\resizebox{\columnwidth}{!}{
\begin{tabular}{lcccc}
\toprule
\textbf{Method} & \textbf{Type} & \textbf{Params} & \textbf{GFLOPs/frame} & \textbf{FPS} \\
\midrule
AVISM / ACVIS & End-to-end & 238M & $\sim$308 & $\sim$2--5 \\
AVTracker    & Pipeline   & $\sim$9.4B & $\sim$18{,}900 & 0.21 \\
\bottomrule
\end{tabular}
}
\end{table}
 
\begin{table}[h]
\centering
\caption{Per-component FLOPs breakdown of AVTracker.}
\label{tab:flops_breakdown}
\resizebox{\columnwidth}{!}{
\begin{tabular}{lcc}
\toprule
\textbf{Component} & \textbf{GFLOPs/frame} & \textbf{Proportion} \\
\midrule
Whisper                  & $\sim$35      & $<$1\% \\
SAM3-Video               & $\sim$630     & 3\% \\
Qwen3-VL-8B-Instruct     & $\sim$18{,}200 & 97\% \\
\bottomrule
\end{tabular}
}
\end{table}
 
Qwen3-VL dominates the cost. Despite being roughly $60\times$ more expensive than end-to-end baselines, AVTracker delivers substantially higher HOTA, suggesting that for the difficulty level of AVTrack the foundation-model orchestration paradigm is currently the more accuracy-efficient choice. Quantization, batched API serving, and KV-cache reuse across the Local and Global Reasoners are natural avenues for reducing this overhead.

% \section{Further Analyses and Discussion}
% \label{sec:further_analyses}
 
% In this section, we provide additional implementation details, complexity and sensitivity analyses, fine-grained per-challenge results, and extended discussion on the role of each modality in our framework. Throughout, the main goal is to clarify the design rationale of AVTracker and to characterize the difficulty of AVTrack from multiple angles.

\section{Per-Challenge Performance Analysis}
\label{sec:per_challenge}
 
To localize where current methods struggle, we report HOTA broken down by the eight challenge categories of AVTrack in \cref{tab:per_challenge}. The category proportions in AVTrack are: Camera Motion $90.5\%$, Visual Occlusion $80.9\%$, Position Change $70.7\%$, Background Switch $60.5\%$, Multi-turn Sounding $56.8\%$, Instance Scale Dynamics $56.9\%$, and Audio--Visual Inconsistency $9.2\%$. Each of these proportions far exceeds those in prior benchmarks such as AVISeg, where most categories appear in less than $10\%$ of clips.
 
\begin{table}[h]
\centering
\caption{Per-challenge HOTA comparison across all baselines.}
\label{tab:per_challenge}
\small
\resizebox{\columnwidth}{!}{
\begin{tabular}{lcccc}
\toprule
\textbf{Challenge} & \textbf{AVISM} & \textbf{ACVIS} & \textbf{AVTrackFormer} & \textbf{AVTracker} \\
\midrule
Multiple Instances   & 20.2 & 20.1 & 19.8 & \textbf{27.0} \\
Instance Scale       & 18.6 & 18.2 & 17.8 & \textbf{26.8} \\
Visual Occlusion     & 20.1 & 19.8 & 19.2 & \textbf{28.1} \\
Position Change      & 20.6 & 20.7 & 20.7 & \textbf{25.9} \\
Camera Motion        & 21.0 & 20.9 & 19.9 & \textbf{29.6} \\
Background Switch    & 20.1 & 19.9 & 19.0 & \textbf{28.9} \\
Multi-turn Sound     & 21.1 & 20.8 & 20.1 & \textbf{29.2} \\
AV Inconsistency     & 12.6 & 12.8 & 12.2 & \textbf{18.5} \\
\bottomrule
\end{tabular}
}
\end{table}

\paragraph{Failure modes.}
Within AVTracker, the relatively weaker scenarios are Visual Occlusion (HOTA $28.1$) and Multiple Instances ($27.0$), both well below the strongest Camera Motion case ($29.6$). The dominant failure pattern in these categories is tracklet association ambiguity when occluded speakers alternate at close range, leading the Global Reasoner to merge or split identities incorrectly. Even in these worst-case scenarios, AVTracker still leads the strongest end-to-end baseline by more than $7$ HOTA, suggesting that the local-to-global formulation degrades gracefully under heavy occlusion rather than collapsing.

\section{Comparison with Commercial Omni-LLMs}
\label{sec:commercial_avllm}
 
As shown in \cref{tab:commercial_av_llm}, to position AVTrack against the rapidly improving frontier of commercial audio-visual LLMs, we run Gemini 2.5 Pro in a zero-shot regime: video frames at $1$ FPS and the raw audio track are fed in directly (no transcript), and the model is prompted to output the active-speaker bounding box per frame together with a cross-frame identifier.
Gemini 2.5 Pro reaches only HOTA $= 14.4$, below specifically trained end-to-end baselines ($\sim$20) and far below AVTracker ($29.1$). Despite their breadth on general audio--visual understanding tasks, current commercial AV LLMs still struggle to localize and track active speakers directly under the conditions assembled in AVTrack, which reinforces the value of AVTrack as a testbed for the next generation of these models.
 
\begin{table}[h]
\centering
\caption{Comparison with commercial audio-visual LLMs.}
\label{tab:commercial_av_llm}
\resizebox{\columnwidth}{!}{
\begin{tabular}{lcccc}
\toprule
\textbf{Method} & \textbf{Type} & \textbf{HOTA} & \textbf{DetA} & \textbf{AssA} \\
\midrule
AVISM          & End-to-end trained    & 20.8 & 23.2 & 19.5 \\
ACVIS          & End-to-end trained    & 20.6 & 22.6 & 19.7 \\
Gemini 2.5 Pro & AV LLM (zero-shot)    & 14.4 & 13.8 & 15.9 \\
AVTracker      & Pipeline (training-free) & 29.1 & 31.2 & 28.5 \\
\bottomrule
\end{tabular}
}
\end{table}

\section{Scope and Future Directions}
\label{sec:scope_future}
 
\subsection{Human-centric Focus}
AVTrack focuses on human-centric scenes for three main reasons. First, human understanding is central to many real-world applications, such as video conferencing, surveillance, accessibility, and human-computer interaction. Second, humans introduce challenges that differ from those in generic object tracking, including multi-turn speech, long-term speaker identity association, and subtle appearance differences between individuals. These factors make dedicated evaluation necessary. Third, AVTrack complements existing benchmarks such as AVISeg, which already addresses general audio-visual instance segmentation, by providing a benchmark specifically designed for human-centric scenarios.

\subsection{Future Work}
Looking forward, we plan to extend AVTrack along three concrete directions. First, we will enlarge the annotations to support \emph{human-centric reference audio-visual segmentation} (\textbf{Human-centric Ref-AVS}), in which models must localize and segment specific individuals based on natural language expressions in scenes with rich auditory and visual cues. This extension is intended to encourage multimodal reasoning that integrates auditory, visual, and linguistic information for precise human-centric scene understanding. Second, motivated by the per-challenge analysis in \cref{sec:per_challenge}, we target architectural improvements for the hardest categories, especially Audio-Visual Inconsistency and Instance Scale Dynamics, where all current methods plateau. Third, we will explore agentic reasoning mechanisms such as memory-augmented trackers and reflection-based error correction, which are well matched to the local-to-global formulation and could further unlock VLM reasoning in long and crowded sequences.

%%%%%%%%%%%%%%%%%%%%%%%%%%%%%%%%%%%%%%%%%%%%%%%%%%%%%%%%%%%%%%%%%%%%%%%%%%%%%%%
%%%%%%%%%%%%%%%%%%%%%%%%%%%%%%%%%%%%%%%%%%%%%%%%%%%%%%%%%%%%%%%%%%%%%%%%%%%%%%%

\end{document}